\pdfoutput=1
\documentclass[11pt]{article}

\usepackage{EMNLP2023}

\usepackage{times}
\usepackage{latexsym}

\usepackage[T1]{fontenc}
\usepackage[utf8]{inputenc}

\usepackage{microtype}

\usepackage{inconsolata}

\usepackage{graphicx}
\usepackage{enumerate}
\usepackage[colorinlistoftodos]{todonotes}
\usepackage{booktabs}
\usepackage{multirow}
\usepackage{hyperref}
\usepackage{caption}
\usepackage{subcaption}
\usepackage{amssymb}
\usepackage{array}
\usepackage{nicematrix}
\usepackage{longtable}
\usepackage{ltablex}

\title{Revising with a Backward Glance: Regressions and Skips during Reading as Cognitive Signals for Revision Policies in Incremental Processing}

\author{Brielen Madureira$^{\mathbf{1}}$ \hspace{10mm} Pelin Çelikkol$^{\mathbf{1}}$ \hspace{10mm}  David Schlangen$^{\mathbf{1, 2}}$ \\
$^{\mathbf{1}}$Computational Linguistics, Department of Linguistics \\ University of Potsdam, Germany \\
$^{\mathbf{2}}$German Research Center for Artificial Intelligence (DFKI), Berlin, Germany \\
\texttt{\{madureiralasota,aynur.celikkol,david.schlangen\}@uni-potsdam.de}}

\begin{document}
\maketitle
\begin{abstract}
In NLP, incremental processors produce output in instalments, based on incoming prefixes of the linguistic input. Some tokens trigger revisions, causing edits to the output hypothesis, but little is known about why models revise when they revise. A policy that detects the time steps where revisions should happen can improve efficiency. Still, retrieving a suitable signal to train a revision policy is an open problem, since it is not naturally available in datasets. In this work, we investigate the appropriateness of regressions and skips in human reading eye-tracking data as signals to inform revision policies in incremental sequence labelling. Using generalised mixed-effects models, we find that the probability of regressions and skips by humans can potentially serve as useful predictors for revisions in BiLSTMs and Transformer models, with consistent results for various languages.
\end{abstract}

\section{Introduction}
\label{sec:intro}

``\textit{Supreme court plans an attack on independent judiciary, says Labour}.'' This was the headline of a news article,\footnote{Source: \href{https://www.theguardian.com/law/2020/nov/15/supreme-court-plans-an-attack-on-independent-judicary-says-labour}{The Guardian, Nov 15, 2020}. Retrieved from \href{https://languagelog.ldc.upenn.edu/nll/?p=49212}{the Language Log blog}.} which sounds incongruous until one interprets it the way intended. That is a \textit{crash blossom},\footnote{\url{https://en.wiktionary.org/wiki/crash\_blossom}} a sentence that becomes ambiguous \textit{e.g.}~due to brevity. The correspondent later \textit{revised} the headline to remove the ambiguity. You probably had to go back and read that sentence again. Such movement is called \textit{regression} in the eye-tracking literature, when the eye makes a regressive, as opposed to progressive, saccade while reading a text. 

\begin{figure}[ht!]
    \centering
    \includegraphics[trim={0 8.3cm 2cm 0},clip,width=\linewidth]{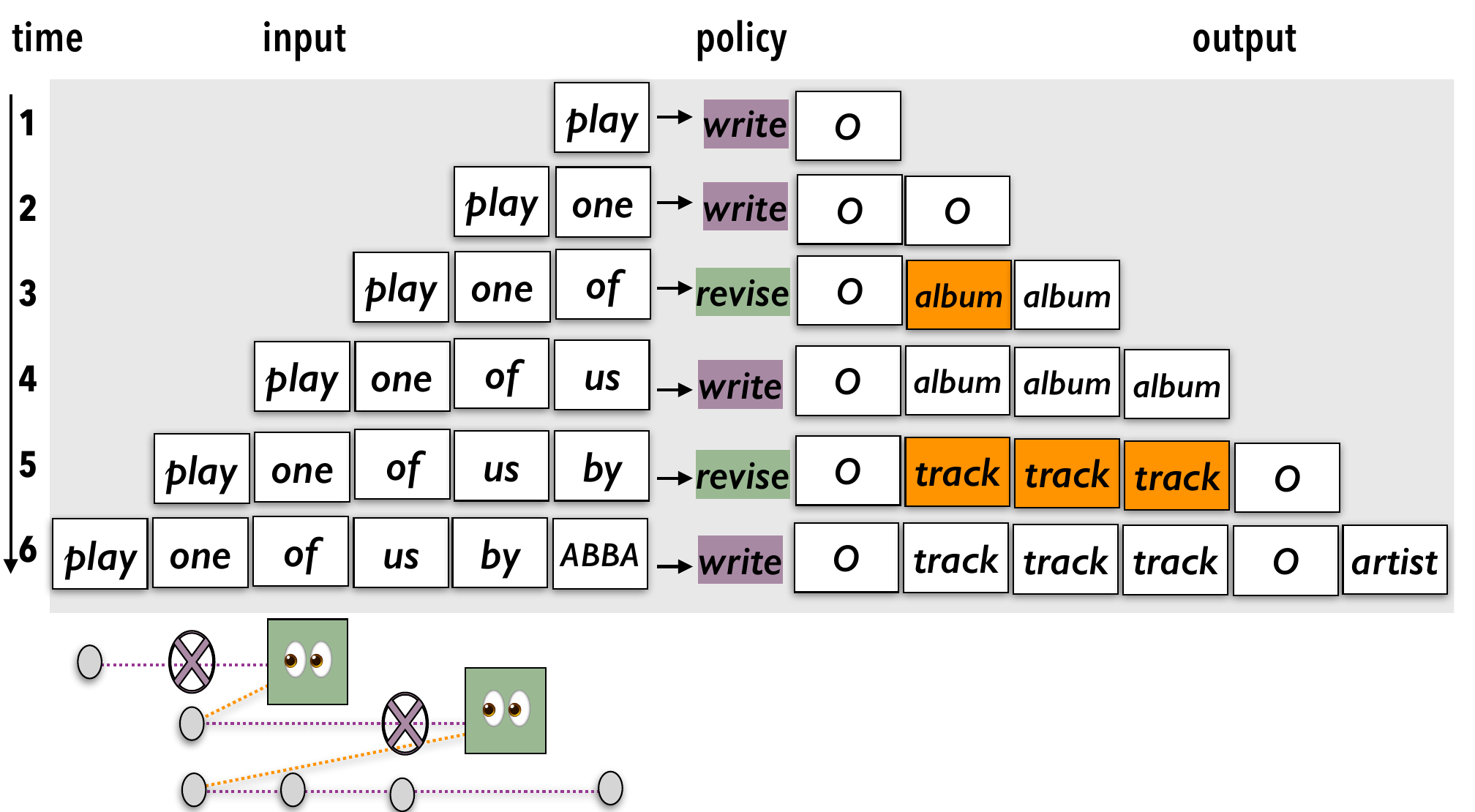}
    \caption{A constructed example of incremental sequence labelling where revisions occur at time steps 3 and 5. If tokens where humans initiate regressions in reading align with tokens that trigger revisions, it can be a cognitive signal to model a revision policy. 
    }
    \label{fig:init-example}
\end{figure}

In incremental NLP models, partial output hypotheses are built at each time step, based on incoming input prefixes, which renders revisability a desirable property to correct mistakes \citep{iu-restart}. This mode takes place in interactive settings that require real-time processing, for instance disfluency detecion or reference resolution in dialogue \citep{Hough-2015,kennington2017simple} and simultaneous translation \citep{cho2016can,arivazhagan-etal-2020-translation,sen-etal-2023-self}. 

Figure \ref{fig:init-example} depicts a constructed example for sequence labelling. For each new token, the model either just extends the current output prefix with a new label, or also edits the output by changing previous labels (here at time steps 3 and 5). Modelling a policy that predicts when revisions should occur is an open research problem, because this signal is not naturally available in the training data \citep{kohn-2018-incremental,tapir}. Moreover, we currently lack evaluation methods to understand whether the revisions performed by a model are linguistically or cognitively motivated (\textit{i.e.}~being grounded in the linguistic input or resembling cognitive processes) or an idiosyncratic result of its internal processing patterns. 

In eye-tracking experiments, many measures can be extracted per token while humans read texts \citep{rayner1998eye}. Common data formats include variables representing whether each token, in first-pass reading, was skipped, fixated and left progressively or triggered a regressive eye movement. In Figure \ref{fig:init-example}, the constructed scanpath shows regressions at tokens \textit{of} and \textit{by} and skips at \textit{one} and \textit{us}. Various theories exist to account for why humans regress (see $\S$\ref{sec:litreview}), but the fact that underlying cognitive processes cause the eyes to move forward or backward at each word (or skip it) lends itself as a cognitively motivated token-level signal.

In this paper, we bridge the concepts of \textit{revisions} in incremental sequence labelling and \textit{regressions} in human eye-tracking reading data. We investigate whether regressions and skips can aid the prediction of revisions in incremental processors, and conclude that eye-tracking measures are a potential cognitively-motivated learning signal to model revision policies.

\section{Motivation}
\label{sec:motiv}
Currently on-trend models like Bi-LSTMs \citep{schuster1997bidirectional} and Transformers \citep{vaswani2017attention} operate in a non-incremental fashion, relying on the availability of complete input sentences or texts to deliver output. One workaround to employ non-incremental encoders in real-time is applying a restart-incremental interface \citep{iu-restart}, enabling outputs to be revised as a by-product of recomputations, as explored by \citet{madureira-schlangen-2020-incremental} and \citet{kahardipraja-etal-2021-towards}. Although possible, it forces recomputation from scratch at every new piece of input, which increases the computational load and can become infeasible for long sequences \cite{kahardipraja-etal-2021-towards}. On the other hand, inherently incremental models like RNNs have the disadvantage of not being able to recover from mistakes via revisions (at least their prototypical versions). 

The sweet spot would be a model that can detect the need to revise. Initiatives in this direction are \textsc{Hear} \citep{kaushal2023efficient}, which has a module that predicts the need to \textit{restart}, and \textsc{Tapir} \citep{tapir}, which integrates an RNN with a Transformer-revisor, predicting whether to \textit{recompute} or to just extend the current output. A difficulty encountered in the latter is how to obtain a ground-truth signal for the revision policy. They derived silver labels from the outputs of another Transformer, which is possibly too model-specific and its linguistic motivation is not explored. \textsc{Hear} compares partial outputs to the non-incremental gold standard which, however, does not encode locally valid hypotheses (which only future input will rule out) and does not accommodate the fact that the gold standard may differ from its final output, thus penalising the incremental metrics with the model's non-incremental deficits \citep{baumann-incremental,inc-revisions}.

We usually do not have corpora containing annotation for the incremental hypotheses for input prefixes by humans, only the annotated gold labels for the final output. But there is vast literature using human reading data as a supervision signal in NLP tasks \citep{barrett2020sequence,mathias2021survey}. Inspired by that, we ask ourselves whether a model's revisions coincide with human regressions in eye-tracking reading data. A positive answer would mean that human reading data could help modelling a dedicated policy for revisions (as opposed to naive recomputations or restarts), and would serve as a cognitively motivated yardstick to judge a models' revisions. 

Among all revisions, some are \textit{effective}, \textit{i.e.}~they edit the prefix into a better state, with respect to a gold standard or to the final output \citep{inc-revisions}. Identifying them can contribute to reducing undesired revisions, which cause instability without bringing the advantage of improvement in output quality. Therefore, if human reading behaviour can help perform only effective revisions, the signal is even more useful for incremental processing.

\section{Related Literature}
\label{sec:litreview}
 
During reading, humans fixate the gaze on some words and make saccades that can be progressive or regressive with respect to the order of the words in the text, so that scanpaths and various measures regarding gaze position, direction and duration can be extracted with eye-tracking devices \citep{rayner2012psychology}, a technique that is becoming more accessible at scale \citep{ribeiro2023webqamgaze}. 

Research based on eye-tracking reading data often rely on the eye-mind hypothesis, which assumes that the eye remains fixated on a word as long as it is being processed \citep{just1980theory}. 
Various research fields rely on the temporal and spatial dimensions of human reading data. We identify at least three (non-mutually exclusive) uses. A consolidated line of research involves studying human cognition and verifying linguistic theories of sentence processing (\textit{e.g.}~\citet{demberg2008data} and \citet{shain-etal-2016-memory-access}).
Another field is occupied with understanding to what extent computational models like artificial neural networks resemble human cognition in how they process language, for example by estimating their psychometric predictive power \citep{wilcox2020predictive,hollenstein-etal-2021-multilingual}. A relationship commonly investigated is the surprisal of language models \textit{versus} human reading time \citep{fernandez-monsalve-etal-2012-lexical,goodkind-bicknell-2018-predictive,wilcox2020predictive}. NLP has been incorporating eye-tracking data in recent years \citep{iida-etal-2013-investigation,tokunaga-etal-2017-eye}, with the emerging use of human reading data both as input to enhance NLP models (
see \citet{barrett2020sequence} and \citet{mathias2021survey} for recent surveys) and as a means for their interpretability \cite{tokunada-interpret-2023}.

In this work, the phenomenon of interest is \textit{regressions}, \textit{i.e.}~eye movements that move backwards in the text and can be shorter or longer-range \citep{rayner2012psychology}. They are a common topic in psycholinguistics research \citep{paape2022reanalysis, paape2021does} and various hypotheses account for their role, such as comprehension or word identification difficulties, low-level visuomotor processes, rereading, memory cues and tools for language processing (see \citet{vitu2005visual}, \citet{lopopolo-etal-2019-dependency} and \citet{booth2013function} for comprehensive discussions and references). Relevant measures are at which word a regression initiates, at which word it lands, regression path duration (how long the reader remains in past text before progressing to unseen text), and how many regressions are initiated for each word. We can also differentiate between first-pass 
and subsequent regressions. 

\paragraph{\textbf{Regressions in NLP}} Reading data has been used as a source of psycholinguistic information for various NLP tasks. When it comes to regressions, \citet{barrett-sogaard-2015-reading} used eye-movements to predict syntactic categories, an idea further explored in \citet{barrett-etal-2016-weakly}, who augmented PoS-taggers with various gaze features, among which was the number of regressions originating from a word. \citet{barrett-sogaard-2015-using} used the number of regressions from and to a word as features to predict grammatical functions. The number of total regressions per word was also used as a feature by \citet{mishra2016predicting} for sarcasm understandability prediction. Regression duration, \textit{i.e.}~the total time spent on a word after the first pass over it, was a useful feature for sentence compression proposed by \citet{klerke-etal-2016-improving}. 
Regressions during coreference resolution annotation were investigated by \citet{cheri-etal-2016-leveraging}, who used it to propose a heuristic for pruning candidates in a coreference resolution model. In \citet{hollenstein-zhang-2019-entity}, the total duration of regressions from a word was used as a context feature in named-entity recognition. 

We draw inspiration from the work by \citet{lopopolo-etal-2019-dependency}, who hypothesised that backward saccades are involved in online syntactic analysis, in which case regressions should coincide, at least partially, with the edges of the relations computed by a dependency parser. They found a significant effect of the number of left-hand side dependency relations on the number of backward saccades. While the authors were interested at predicting human regressions from a model instantiating a parsing theory, we are conversely interested in using human regressions as a signal to train an NLP model.\footnote{It is also worth investigating whether a model's revisions can predict human regression behaviour, but it is beyond the scope of this work.}

\begin{figure*}[ht]
    \centering
    \includegraphics[trim={0 16.5cm 0cm 0},clip,width=\linewidth]{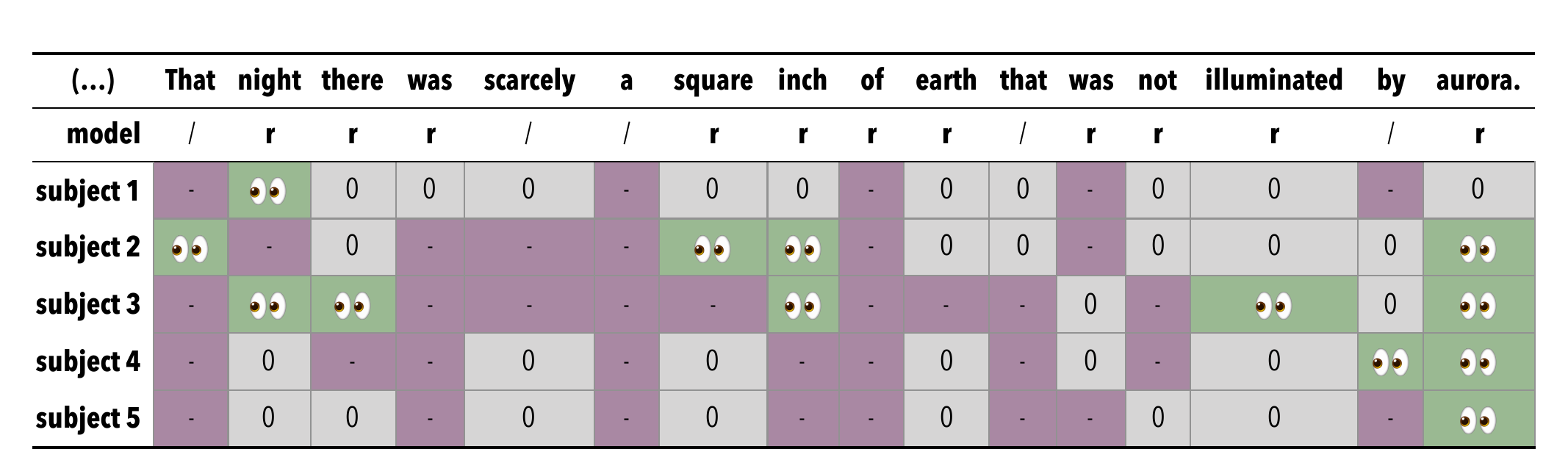}
    \caption{An example of our data structure for a portion of a text in the Provo corpus, processed by a restart-incremental Transformer predicting dependency relations. Each token is annotated with the reading variable for each subject (eyes: regressed, 0: not regressed, -: skipped) and the model's decision (r: revised, /: not revised).}
    \label{fig:annotation-example}
\end{figure*}

\section{Method}
\label{sec:methods}
To perform the analysis, we use binomial generalised linear mixed models (GLMM) with a logit link function to predict model revisions. Similar to the approach by \citet{lopopolo-etal-2019-dependency}, for each combination of dataset and NLP model/task, we fit two GLMMs: The baseline model~(\ref{eqn:null}) only includes the token position variable as a fixed effect and texts as random effects. Since a model's revisions may vary depending on the word's position in the text, we add token position as a baseline predictor and include texts to account for any variability due to different types of texts. We fit model~(\ref{eqn:full}) with the same structure, adding the predictors of regression probability and skipping probability as fixed effects. 
The binary dependent variable is a token's revise/not-revise label.

\begin{equation}
\label{eqn:null}
    \begin{aligned}
    model\;revision & \sim  token\;position & & & & & \\
                    & + (1|text) & & & & &
    \end{aligned}
\end{equation}

\begin{equation}
\label{eqn:full}
    \begin{aligned}
    model\;revision & \sim  token\;position & & & & &\\
                    & + p(regression) & & & & &\\ 
                    & + p(skip)  & & & & &\\
                    & + (1|text) & & & & &
    \end{aligned}
\end{equation}

We use likelihood ratio tests (LRT) between the null and the full models to evaluate the goodness of fit. LRTs are used to compare a baseline model to a more complex one with more predictors and decide if certain predictors should be included, consequently selecting the model that fits the data better. To infer statistical significance, we obtain $p$-values using the $\chi^2$ distribution.

We do not intend to make claims about \textit{why} regressions occur. For our purposes, we take at face value that they \textit{did} occur in the eye-tracking experiments (and when). We are interested in words at which regressions are initiated when they are first read, 
knowing that, for some reason, the reader went to past input before continuing (as a consequence, we also analyse words that are not fixated in the first pass). Still, the hypothesis that regressions occur due to reanalysis, when humans encounter garden path sentences \citep{altmann1992avoiding}, is at our favour, since revisions represent updates in the current model's interpretation caused by input seen for the first time.

\section{Data}
\label{sec:data}
In this section, we explain the data structure constructed for the analysis. We then introduce the eye-tracking corpora and the models selected for this study, and discuss how we extract the incremental outputs from non-incremental, pre-trained sequence labelling models.\footnote{The pre-processing scripts and implementation code is available at \url{https://github.com/briemadu/revreg}.}

\paragraph{\textbf{Procedure}} Our method requires knowing, for each token $w$ in a text, what was the behaviour of the model while performing sequence labelling and of the humans while reading the text. More specifically, we need to know whether the model revised its hypothesis upon processing $w$ and whether humans skipped $w$, fixated it but moved forward, or fixated it and regressed. We thus construct an annotation mapping tokens to human and model data as illustrated in Figure \ref{fig:annotation-example}. The texts come from the eye-tracking corpora, from which we also extract the human skips or regressions. The revisions are retrieved by feeding the same texts to the NLP models, prefix by prefix in a restart-incremental fashion, and checking if labels change at each time step.

\begin{table}[ht!]
\centering
   
    { \setlength{\tabcolsep}{4pt}
    \small
    
    \begin{tabular}{llccc}
        \toprule
         & language & tokens & texts & subjects \\
        \cmidrule{2-5}
        MECO-L1 & Dutch & 2,231 & 12 & 45 \\
        MECO-L2 & English (L2) & 1,658 & 12 & 538 \\
        Nicenboim & Spanish & 791 & 48 & 71 \\
        PoTeC & German & 1,895 & 12 & 62 \\
        Provo & English & 2,743 & 55 & 84 \\
        RastrOS & Br. Portuguese & 2,494 & 50 & 37 \\
        \bottomrule
    \end{tabular}
}
\caption{Human reading eye-tracking corpora.}
\label{tab:rt-human-stats}
\end{table}

\begin{table*}[ht!]
\centering
   
    { \setlength{\tabcolsep}{4pt}
    \small

\begin{tabular}{llcccccccccccc}
\toprule
 &  & \multicolumn{2}{c}{\textbf{MECO (du)}} & \multicolumn{2}{c}{\textbf{MECO (enl2)}} & \multicolumn{2}{c}{\textbf{Nicenboim (es)}} & \multicolumn{2}{c}{\textbf{PoTeC (de)}} & \multicolumn{2}{c}{\textbf{Provo (en)}} & \multicolumn{2}{c}{\textbf{RastrOS (ptbr)}} \\
 \cmidrule(lr){3-4}  \cmidrule(lr){5-6}  \cmidrule(lr){7-8}  \cmidrule(lr){9-10}  \cmidrule(lr){11-12}  \cmidrule(lr){13-14}
 & & all-r& eff-r & all-r& eff-r & all-r& eff-r & all-r& eff-r & all-r& eff-r & all-r& eff-r \\
\cmidrule(r){1-2} \cmidrule(lr){3-4}  \cmidrule(lr){5-6}  \cmidrule(lr){7-8}  \cmidrule(lr){9-10}  \cmidrule(lr){11-12}  \cmidrule(lr){13-14}

\multirow[t]{3}{*}{\textbf{BiLSTM}} & deprel & 58.45 & 47.20 & 60.74 & 54.52 & 55.75 & 50.32 & 53.56 & 44.27 & 60.99 & 53.70 & 54.01 & 46.75 \\
 & head & 65.76 & 38.32 & 66.95 & 38.60 & 61.31 & 43.36 & 67.28 & 40.37 & 67.92 & 39.30 & 60.34 & 43.70 \\
 & pos & 12.95 & 11.52 & 11.70 & 10.68 & \ \ 6.32 & \ \ 5.44 & 17.89 & 15.51 & 12.65 & 11.27 & 29.19 & 27.11 \\
 
\cmidrule(r){1-2} \cmidrule(lr){3-4}  \cmidrule(lr){5-6}  \cmidrule(lr){7-8}  \cmidrule(lr){9-10}  \cmidrule(lr){11-12}  \cmidrule(lr){13-14}

 \multirow[t]{3}{*}{\textbf{Transformer}} & deprel & 63.92 & 52.44 & 67.97 & 57.66 & 48.93 & 44.37 & 73.67 & 56.36 & 66.68 & 58.77 & 52.81 & 44.23 \\
 & head & 67.55 & 38.01 & 69.06 & 37.21 & 57.27 & 41.47 & 74.56 & 43.38 & 69.30 & 38.46 & 61.39 & 42.98 \\
 & pos & \ \ 9.82 & \ \ 6.28 & \ \ 7.84 & \ \ 6.09 & \ \ 1.90 & \ \ 1.64 & \ \ 5.01 & \ \ 4.12 & \ \ 8.09 & \ \ 6.56 & \ \ 9.22 & \ \ 7.62 \\
\bottomrule
\end{tabular}

}
\caption{\% of timesteps that trigger revisions (all-r) and effective revisions (eff-r) for each model and task.}
\label{tab:model-stats}
\end{table*}

\begin{figure}[!ht]
    \centering
        \includegraphics[width=\columnwidth]{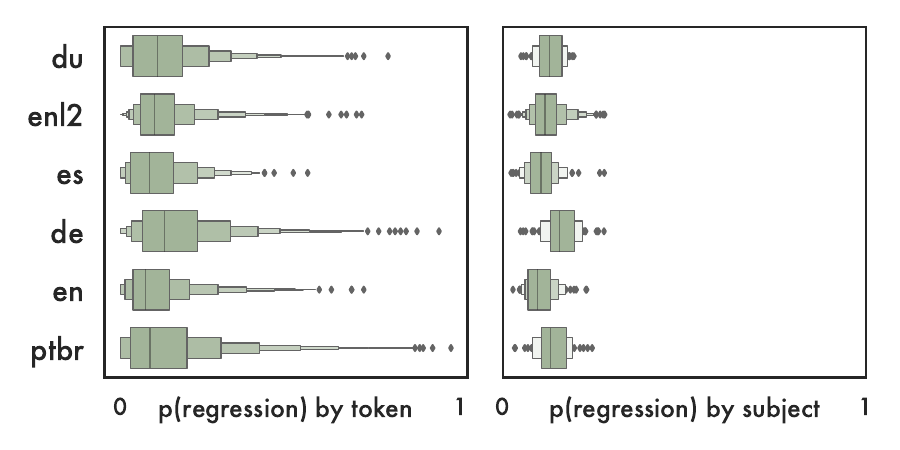}
        \includegraphics[width=\columnwidth]{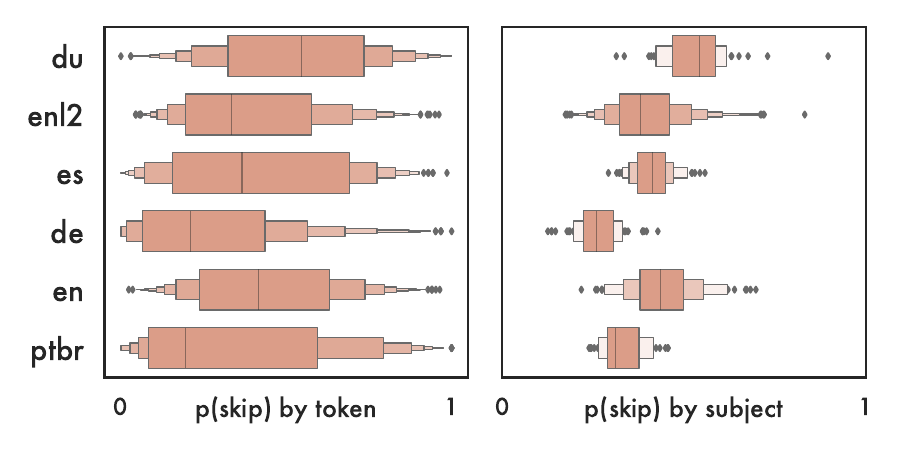}
    \caption{Distributions of the probabilities of regression and skips, by token (left) and by subject (right) estimated from the human reading data for each dataset.}
    \label{fig:human-data-dist}
\end{figure}

\paragraph{\textbf{Human Regressions}} We analyse six eye-tracking human reading corpora: MECO-L1 \citep{siegelman2022expanding}, MECO-L2 \citep{kuperman2023text}, Nicenboim (no official name) \citep{nicenboim2015working}, PoTeC \citep{makowski2019discriminative,jager2020deep}, Provo \citep{luke2018provo} and RastrOS \citep{vieira-rastros,leal2022rastros}. Table \ref{tab:rt-human-stats} presents their language and size. The distribution of regressions and skips (per token and per subject) is shown in Figure \ref{fig:human-data-dist}. Although many other corpora exist, we opted to use those that had first-pass regression and first-pass skip measures already available or easy to infer from other measures. For each interest area,\footnote{An interest area sometimes includes more than one token, \textit{e.g}~a word and punctuation, like \textit{aurora.} in Figure \ref{fig:annotation-example}.} we retrieve the label for each subject as follows: If the token was skipped in the first-pass reading, we label it as skipped. Otherwise, we retrieve a variable which is 1 if a first-pass regression was initiated at that interest area, and 0 otherwise. Although regressions can occur later, we only consider what happens in the first-pass reading to approximate what the model does (revisions happen when a token is integrated for the first time in the sequence). The probabilities are estimated by computing the proportion of regressions and skips per token (excluding subjects with missing data), following existing literature in terms of using average human behaviour as a feature \citep{barrett-etal-2016-weakly,hollenstein-zhang-2019-entity}. We checked that they are only moderately (negatively) correlated ($-0.59<\rho<-0.44$). See Appendix for details about the measures and pre-processing.

\paragraph{\textbf{Models' Revisions}} We opt to evaluate pre-trained sequence labelling models with a restart-incremental paradigm. Models were selected according to the availability of languages to match the eye-tracking corpora. We evaluate Stanza's BiLSTM models \citep{qi-etal-2020-stanza}\footnote{\url{https://github.com/stanfordnlp/stanza}.} and Explosion's pre-trained multi-task Transformer architectures.\footnote{Release documented in \url{https://explosion.ai/blog/ud-benchmarks-v3-2} and available at their model hub on Hugging Face \url{https://huggingface.co/explosion}.} These families of models were selected due to the availability of all languages and comparability in terms of similar training data, as both were trained on the Universal Dependencies corpora \citep{ud-project}. The model checkpoints for each language are listed in Table \ref{tab:pretrained-models}. We extract the incremental outputs for dependency parsing (prediction of the head position and the relation) and POS-tagging. We also inspected NER, but revisions were extremely sparse in these datasets (possibly due to the genres of the texts), so we did not analyse it further. The same texts from the eye-tracking data are fed to each model, one prefix after another, as illustrated in Figure \ref{fig:init-example}, following previous works \citep{madureira-schlangen-2020-incremental,kahardipraja-etal-2021-towards}. At each time step, we extend the input with one interest area (\textit{i.e.}, sometimes it means more than one token). If the output prefix at time $t$ (apart from the recently added label(s), which refer to the last interest area) differs from the output at time $t-1$, a revision occurred. If more labels match the final output than in the previous prefix, the revision is effective. The percentage of (effective) revisions over tokens/timesteps is shown in Table \ref{tab:model-stats}.

\begin{table}[ht!]
\centering
   
    { \setlength{\tabcolsep}{4pt}
    \small
    
    \begin{tabular}{lll}
        \toprule
         & Explosion & Stanza \\
        \midrule
        MECO-L1 & \texttt{nl\_udv25\_dutchalpino\_trf} & nl \\
        MECO-L2 & \texttt{en\_udv25\_englishewt\_trf} &  en \\
        Nicenboim & \texttt{es\_udv25\_spanishancora\_trf} & es \\
        PoTeC & \texttt{de\_udv25\_germanhdt\_trf} & de \\
        Provo & \texttt{en\_udv25\_englishewt\_trf} & en \\
        RastrOS & \texttt{pt\_udv25\_portuguesebosque\_trf} & pt \\
        \bottomrule
    \end{tabular}
}
\caption{Specification of the pre-trained NLP models.}
\label{tab:pretrained-models}
\end{table}

\begin{table*}[ht!]
\centering
   
    { \setlength{\tabcolsep}{4pt}
    \small

\begin{tabular}{lllrrrrrrrr}
\toprule
\midrule
 &  &  & \multicolumn{2}{c}{\textbf{estimate}} & \multicolumn{2}{c}{\textbf{SE}} & \multicolumn{2}{c}{\textbf{z}} & \multicolumn{2}{c}{\textbf{p}} \\
 \cmidrule(lr){4-5}\cmidrule(lr){6-7}\cmidrule(lr){8-9}\cmidrule(lr){10-11}
 &  & & \multicolumn{1}{c}{MECO-L2} & \multicolumn{1}{c}{Provo} & \multicolumn{1}{c}{MECO-L2} & Provo & \multicolumn{1}{c}{MECO-L2} & \multicolumn{1}{c}{Provo} & \multicolumn{1}{c}{MECO-L2} & Provo \\
\cmidrule(r){1-3}\cmidrule(lr){4-5}\cmidrule(lr){6-7}\cmidrule(lr){8-9}\cmidrule(lr){10-11}
\multirow[t]{12}{*}{\textbf{BiLSTM}} & \multirow[t]{4}{*}{deprel} & intercept & 1.29*** & 1.22*** & 0.05 & 0.05 & 24.18 & 24.29 & <0.001 & <0.001 \\
 &  & p(reg) & 3.41*** & 3.30*** & 0.05 & 0.09 & 73.39 & 38.56 & <0.001 & <0.001 \\
 &  & p(skip) & -2.80*** & -3.68*** & 0.02 & 0.03 & -178.47 & -133.52 & <0.001 & <0.001 \\
 &  & position & -0.03*** & 0.21*** & 0.00 & 0.01 & -8.94 & 38.87 & <0.001 & <0.001 \\
\cmidrule(r){2-3}\cmidrule(lr){4-5}\cmidrule(lr){6-7}\cmidrule(lr){8-9}\cmidrule(lr){10-11}
 & \multirow[t]{4}{*}{head} & intercept & 1.59*** & 1.76*** & 0.06 & 0.05 & 27.44 & 33.12 & <0.001 & <0.001 \\
 &  & p(reg) & 4.32*** & 2.18*** & 0.05 & 0.10 & 81.05 & 21.84 & <0.001 & <0.001 \\
 &  & p(skip) & -3.23*** & -4.92*** & 0.02 & 0.03 & -193.35 & -155.18 & <0.001 & <0.001 \\
 &  & position & - & 0.40*** & - & 0.01 & - & 68.85 & - & <0.001 \\
\cmidrule(r){2-3}\cmidrule(lr){4-5}\cmidrule(lr){6-7}\cmidrule(lr){8-9}\cmidrule(lr){10-11}
 & \multirow[t]{4}{*}{pos} & intercept & -2.62*** & -1.92*** & 0.07 & 0.08 & -36.21 & -22.77 & <0.001 & <0.001 \\
 &  & p(reg) & 1.25*** & 1.42*** & 0.05 & 0.08 & 27.53 & 18.61 & <0.001 & <0.001 \\
 &  & p(skip) & -1.16*** & -0.66*** & 0.02 & 0.04 & -52.26 & -18.63 & <0.001 & <0.001 \\
 &  & position & 0.20*** & - & 0.00 & - & 42.18 & - & <0.001 & - \\
\cmidrule(r){1-3}\cmidrule(lr){4-5}\cmidrule(lr){6-7}\cmidrule(lr){8-9}\cmidrule(lr){10-11}
\multirow[t]{12}{*}{\textbf{Transformer}} & \multirow[t]{4}{*}{deprel} & intercept & 1.22*** & 1.28*** & 0.09 & 0.05 & 14.28 & 24.39 & <0.001 & <0.001 \\
 &  & p(reg) & 4.39*** & 3.26*** & 0.05 & 0.09 & 82.91 & 34.39 & <0.001 & <0.001 \\
 &  & p(skip) & -2.53*** & -3.75*** & 0.02 & 0.03 & -154.71 & -129.34 & <0.001 & <0.001 \\
 &  & position & 0.03*** & 0.30*** & 0.00 & 0.01 & 11.37 & 54.95 & <0.001 & <0.001 \\
\cmidrule(r){2-3}\cmidrule(lr){4-5}\cmidrule(lr){6-7}\cmidrule(lr){8-9}\cmidrule(lr){10-11}
 & \multirow[t]{4}{*}{head} & intercept & 1.45*** & 1.45*** & 0.08 & 0.05 & 18.13 & 29.17 & <0.001 & <0.001 \\
 &  & p(reg) & 4.40*** & 2.27*** & 0.05 & 0.10 & 82.24 & 23.76 & <0.001 & <0.001 \\
 &  & p(skip) & -2.64*** & -4.01*** & 0.02 & 0.03 & -160.14 & -133.24 & <0.001 & <0.001 \\
 &  & position & - & 0.37*** & - & 0.01 & - & 64.92 & - & <0.001 \\
\cmidrule(r){2-3}\cmidrule(lr){4-5}\cmidrule(lr){6-7}\cmidrule(lr){8-9}\cmidrule(lr){10-11}
 & \multirow[t]{4}{*}{pos} & intercept & -2.64*** & -2.69*** & 0.17 & 0.14 & -15.28 & -19.71 & <0.001 & <0.001 \\
 &  & p(reg) & -0.62*** & 3.00*** & 0.06 & 0.10 & -9.49 & 31.11 & <0.001 & <0.001 \\
 &  & p(skip) & -0.77*** & 0.80*** & 0.03 & 0.04 & -29.33 & 18.07 & <0.001 & <0.001 \\
 &  & position & 0.08*** & -0.25*** & 0.01 & 0.01 & 15.56 & -30.18 & <0.001 & <0.001 \\
\midrule
\bottomrule
\end{tabular}

}
\caption{Overview of the GLMM results, showing the estimated coefficients for each variable and their statistical significance, for the English corpora. See Appendix for the the complete table.}
\label{tab:results-short}
\end{table*}

\section{Results}
\label{sec:analysis}
We summarise the full GLMM results in Table~\ref{tab:results-short} for Provo and MECO-L2 datasets. Due to a large number of experiments, we only present results for the English models in this table; the complete results are in the Appendix. In every (dataset, NLP model, task) combination, the likelihood ratio test between the baseline and full models revealed that the full model, including the two predictors of interest, is a better fit to the data than the baseline model with only token position and text. 

\begin{figure*}[t]
    \centering
    \includegraphics[trim={0 0cm 0cm 0},clip,width=\linewidth]{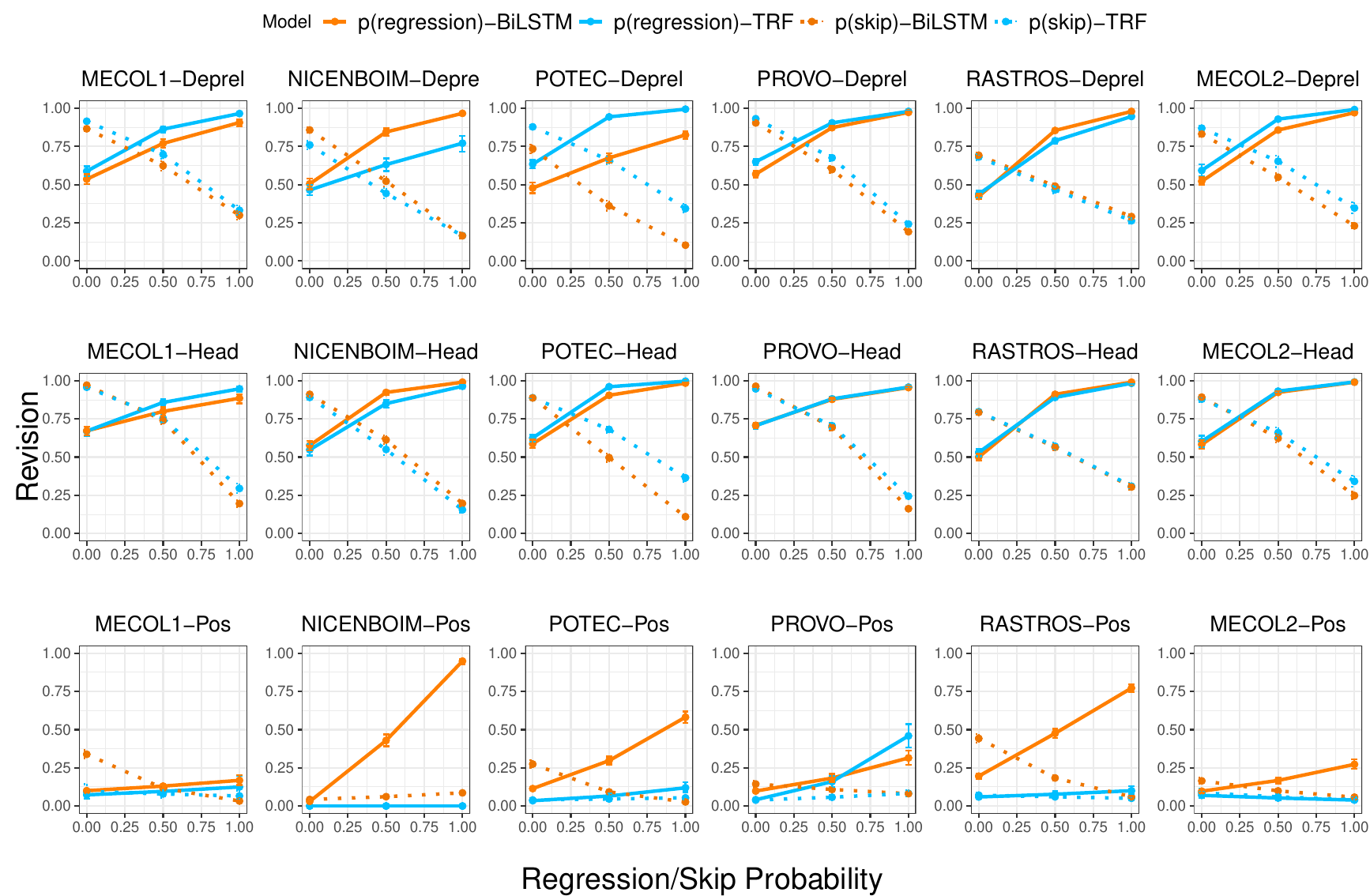}
    \caption{The full GLMM predictions of the revision probability are shown. Each plot presents the predictions for BiLSTM and Transformer models given regression and skip probability in the corresponding dataset. Error bars represent 95\% confidence interval.}
    \label{fig:pred-plots}
\end{figure*}

The token position was a significant predictor of revisions in most models. For the few cases in which it did not significantly affect revisions (\textit{i.e.}, MECO-L2-Transformer-head and BiLSTM-head, MECO-L1-BiLSTM-head, Provo-BiLSTM-pos), we fitted models without this predictor instead. 

We found that average human gaze patterns, namely the estimated word's regression and skip probability, were significant predictors of revisions. This was a consistent result across all eye-tracking corpora, for the BiLSTM and the Transformer, both for dependency parsing and POS-tagging.
On the one hand, human regressions were often positively related to revisions, so that words with a higher regression probability were more likely to be revised by models (MECO-L2-Transformer-pos was the only exception where regression probability negatively affected revisions). Conversely, a word's skip probability decreased the probability of it triggering a revision in most cases (with the exceptions of Potec and Provo-Transformer-pos and Nicenboim-BiLSTM-pos). These relationships are illustrated in Figure~\ref{fig:pred-plots}. The magnitude of the regression coefficient did not follow a general pattern for the tasks, but the skip coefficient was more often larger for the task of predicting the head than for the dependency relation, which was usually larger than for POS-tagging (exceptions to this is RastrOS-Transformer and MECO-L1-BiLSTM).

In a further analysis, we repeated the same procedure to predict only the effective revisions and observed the same trend in regression and skip coefficients when predicting effective revisions, in terms of direction and significance, in all experiments. However, the magnitude of the coefficients differed, sometimes being larger in one or the other, which does not allow us to draw general conclusions at this point. The coefficient of token position was, in most cases, smaller in the model that predicts effective revisions. Similarly, in many models the magnitude of the coefficient of skips was larger for models predicting effective revisions.

To assess the fit of the model to the data in more detail, we evaluated its predictions by running permutation tests with the null hypothesis that the probabilities assigned to (effective) revisions and to not-revisions are randomly sampled from the same distribution. Besides, we computed the area under the ROC curve in each model. As we can see in Table \ref{tab:preds}, most of the differences were significant (except for many cases in POS-tagging), but their magnitude was relatively small. The AUC was around 0.7 for all datasets, and in some experiments the models of effective revisions had higher AUC. Examples with considerable improvements are RastrOS-head and Nicenboim-head.

\section{Do models revise when humans regress?}
\label{sec:discussion}
We have gathered evidence that there is a relationship between NLP restart-incremental models' revisions and human gaze behaviour in reading, which manifests as the probability of revision at a given token being partially predictable from it being often skipped or triggering regressions, when token position and text are accounted for. Interestingly, the overall findings hold for BiLSTM and Transformers, even though their encoding mechanisms are different, and also for all five languages, despite the eye-tracking data having been collected from different text genres and the readers having performed different tasks (or no additional task beyond reading for comprehension, as in Provo). 

For this conclusion, we did not rely on any assumptions for the connection between human regressions and incremental models' revisions beyond the analogy of what we factually know: When seeing text areas for the first time, humans made decisions to skip or fixate, and possibly to revisit past text, and at some words, models ``decided'' to revisit past decisions. 

\begin{table}[ht!]
\centering
   
    { \setlength{\tabcolsep}{3pt}
    \small

\begin{tabular}{lllllrr}
\toprule
\midrule
 &  &  & \multicolumn{2}{c}{\textbf{abs. mean diff}} & \multicolumn{2}{c}{\textbf{AUC}} \\
\cmidrule(lr){4-5}\cmidrule(lr){6-7}
 &  & & all-r & eff-r & all-r & eff-r \\
\cmidrule(r){1-3}\cmidrule(lr){4-5}\cmidrule(lr){6-7}
\multirow[t]{6}{*}{\textbf{MECO-L1}} & \multirow[t]{2}{*}{deprel} & BiLSTM & 0.13* & 0.16* & 0.71 & 0.74 \\
 &  & Trfmer & 0.15* & 0.14* & 0.73 & 0.72 \\
\cmidrule(r){2-3}\cmidrule(lr){4-5}\cmidrule(lr){6-7}
 & \multirow[t]{2}{*}{head} & BiLSTM & 0.22* & 0.26* & 0.78 & 0.80 \\
 &  & Trfmer & 0.18* & 0.21* & 0.76 & 0.77 \\
\cmidrule(r){2-3}\cmidrule(lr){4-5}\cmidrule(lr){6-7}
 & \multirow[t]{2}{*}{pos} & BiLSTM & 0.05* & 0.05* & 0.69 & 0.71 \\
 &  & Trfmer & 0.03* & 0.02 & 0.68 & 0.66 \\
\cmidrule(r){1-3}\cmidrule(lr){4-5}\cmidrule(lr){6-7}
\multirow[t]{6}{*}{\textbf{MECO-L2}} & \multirow[t]{2}{*}{deprel} & BiLSTM & 0.12* & 0.12* & 0.70 & 0.69 \\
 &  & Trfmer & 0.14* & 0.10* & 0.72 & 0.68 \\
\cmidrule(r){2-3}\cmidrule(lr){4-5}\cmidrule(lr){6-7}
 & \multirow[t]{2}{*}{head} & BiLSTM & 0.15* & 0.20* & 0.73 & 0.76 \\
 &  & Trfmer & 0.12* & 0.22* & 0.70 & 0.77 \\
\cmidrule(r){2-3}\cmidrule(lr){4-5}\cmidrule(lr){6-7}
 & \multirow[t]{2}{*}{pos} & BiLSTM & 0.02* & 0.02* & 0.63 & 0.62 \\
 &  & Trfmer & 0.03* & 0.01* & 0.67 & 0.64 \\
\cmidrule(r){1-3}\cmidrule(lr){4-5}\cmidrule(lr){6-7}
\multirow[t]{6}{*}{\textbf{Nicenboim}} & \multirow[t]{2}{*}{deprel} & BiLSTM & 0.27* & 0.28* & 0.79 & 0.80 \\
 &  & Trfmer & 0.19* & 0.19* & 0.74 & 0.74 \\
\cmidrule(r){2-3}\cmidrule(lr){4-5}\cmidrule(lr){6-7}
 & \multirow[t]{2}{*}{head} & BiLSTM & 0.31* & 0.45* & 0.81 & 0.88 \\
 &  & Trfmer & 0.31* & 0.41* & 0.81 & 0.87 \\
\cmidrule(r){2-3}\cmidrule(lr){4-5}\cmidrule(lr){6-7}
 & \multirow[t]{2}{*}{pos} & BiLSTM & 0.03* & 0.04* & 0.69 & 0.73 \\
 &  & Trfmer & 0.06 & 0.06 & 0.89 & 0.89 \\
\cmidrule(r){1-3}\cmidrule(lr){4-5}\cmidrule(lr){6-7}
\multirow[t]{6}{*}{\textbf{PoTeC}} & \multirow[t]{2}{*}{deprel} & BiLSTM & 0.14* & 0.12* & 0.71 & 0.70 \\
 &  & Trfmer & 0.14* & 0.11* & 0.74 & 0.69 \\
\cmidrule(r){2-3}\cmidrule(lr){4-5}\cmidrule(lr){6-7}
 & \multirow[t]{2}{*}{head} & BiLSTM & 0.23* & 0.28* & 0.79 & 0.81 \\
 &  & Trfmer & 0.15* & 0.22* & 0.75 & 0.77 \\
\cmidrule(r){2-3}\cmidrule(lr){4-5}\cmidrule(lr){6-7}
 & \multirow[t]{2}{*}{pos} & BiLSTM & 0.08* & 0.08* & 0.70 & 0.71 \\
 &  & Trfmer & 0.01 & 0.00 & 0.62 & 0.61 \\
\cmidrule(r){1-3}\cmidrule(lr){4-5}\cmidrule(lr){6-7}
\multirow[t]{6}{*}{\textbf{Provo}} & \multirow[t]{2}{*}{deprel} & BiLSTM & 0.20* & 0.19* & 0.76 & 0.75 \\
 &  & Trfmer & 0.20* & 0.17* & 0.76 & 0.74 \\
\cmidrule(r){2-3}\cmidrule(lr){4-5}\cmidrule(lr){6-7}
 & \multirow[t]{2}{*}{head} & BiLSTM & 0.25* & 0.21* & 0.79 & 0.77 \\
 &  & Trfmer & 0.20* & 0.22* & 0.76 & 0.77 \\
\cmidrule(r){2-3}\cmidrule(lr){4-5}\cmidrule(lr){6-7}
 & \multirow[t]{2}{*}{pos} & BiLSTM & 0.02 & 0.01 & 0.64 & 0.64 \\
 &  & Trfmer & 0.04 & 0.02 & 0.72 & 0.70 \\
\cmidrule(r){1-3}\cmidrule(lr){4-5}\cmidrule(lr){6-7}
\multirow[t]{6}{*}{\textbf{RastrOS}} & \multirow[t]{2}{*}{deprel} & BiLSTM & 0.17* & 0.18* & 0.74 & 0.74 \\
 &  & Trfmer & 0.16* & 0.16* & 0.73 & 0.74 \\
\cmidrule(r){2-3}\cmidrule(lr){4-5}\cmidrule(lr){6-7}
 & \multirow[t]{2}{*}{head} & BiLSTM & 0.22* & 0.32* & 0.77 & 0.83 \\
 &  & Trfmer & 0.21* & 0.31* & 0.76 & 0.82 \\
\cmidrule(r){2-3}\cmidrule(lr){4-5}\cmidrule(lr){6-7}
 & \multirow[t]{2}{*}{pos} & BiLSTM & 0.16* & 0.17* & 0.76 & 0.76 \\
 &  & Trfmer & 0.05* & 0.02 & 0.71 & 0.68 \\
\midrule
\bottomrule
\end{tabular}

}
\caption{Left block: Absolute difference of sample means in the predictions of the models between time steps with and without revisions. * means $p$-value < 0.001. Right block: Area Under the ROC Curve when the fitted models' predictions are used for binary classification of revision time steps in the data.}
\label{tab:preds}
\end{table}

Some exceptions to the general trend in predicting model revisions occurred in POS-tagging, for which relatively fewer revisions occur (see Table 2). The sparsity of revisions may cause the signal to be harder to model well without more data. For dependency parsing, more revisions are expected, especially because in the beginning of the sentence the model has to wait until the root is processed to make good predictions. There may also be a difference in processing, since the humans could regress to previous sentences in the text, whereas the NLP models depend on their internal tokenisation and sentence boundary detection. 

This suggests that eye-tracking measures can be transformed into a useful signal to inform the decision of when to revise in mixed restart-incremental processors, especially when the model's task entails more syntactic tasks with frequent revisions to the input.

Still, preliminary investigation of the revision probabilities predicted by the model did not yield a straightforward threshold for binary classification, despite the difference in means being statistically significant. This invites a more detailed extrinsic evaluation, by incorporating the human predictors into a revision controller like TAPIR \citep{tapir}, and assessing the revisions with the evaluation methods discussed by \citet{inc-revisions}. One approach is to train an incremental sequence labelling model whose revision policy relies on eye-tracking data as part of the input and comparing its performance against a model without it. Since skips had a negative effect, it may also be possible to use other variables that relate to the probability of a token being skipped, like POS-tags or word frequency and length, as additional input, which are cheaper to obtain. The analysis should also be done with larger datasets and other models and tasks.

The usefulness of our findings presupposes the availability of eye-tracking measures during inference on truly unseen data, which is an open problem because such signal is not always available in real time. One possibility is to use pretrained eye-tracking models to predict regressions and skips, as in approaches discussed in the literature \citep{engbert2005swift,eyettention}.

Down the road, a revision policy should not only detect times to revise, but times to revise \textit{effectively}, since wrong revisions make the partial outputs less reliable for downstream processors. Our experiments showed that regressions and skips are also good predictors for effective revisions. Identifying ways to filter this more specific signal demands further investigation. An immediate next step is to evaluate the predictions of each model in unseen data for all revisions and for effective revisions.

\section{Conclusion}
\label{sec:conclusion}
Let us conclude with a \textit{backward glance} to our contribution. We have addressed the open question of whether pre-trained sequence labelling models, when employed incrementally, perform revisions in a similar fashion as humans skip words or make regressive eye movements while reading. We have found a significant effect in all the experiments, supporting the use of human reading data as a cognitive signal to inform revision policies. This is a valuable finding: BiLSTMs and Transformers are bidirectional, trained on full sequences, but if we make them process linguistic input incrementally, their revisions can be partially predicted by human reading behaviour. This is also a step forward towards understanding why these models change hypotheses at some tokens, when only partial prefixes are available.

Besides advancing the research on eye-tracking-augmented NLP, this study also opens the door to exploring other cognitive perspectives with restart-incremental NLP models. We see a potential to go the other direction and investigate to what extent a ``mixed incrementality'' model (architectures relying on an incremental processor with occasional restarts) would capture the patterns of human gaze in reading, and hence function as a model of that. In this case, revisions would serve as predictors of human regressions, with control variables like word frequency, surprisal and word length. Other possibility for future work is to investigate whether other measures, like number of fixations or regressions \textit{to} a token, are related to the edits per label.

\section*{Limitations}
\label{sec:limitations}
Here we summarise a few known limitations that we have mentioned throughout the text. We have analysed various datasets which differ both in the ways they were collected (the task humans were performing, \textit{e.g.}~only reading or also answering to comprehension questions) as well as the length and genre of the texts. The size of the eye-tracking datasets is, in general, small. Ideally, larger amounts of data are necessary to train a revision policy than what we had available for the analysis. Some preprocessing steps had to be made; in particular, some decisions were necessary on had how to merge tokens and interpret documentation, so that a mapping could be created. This is documented in the Appendix, but alternative ways are also possible. We limited the study to families of pre-trained models and tasks for which all languages were available. There can be a mismatch between the humans having the full text available at any point and the models performing sentence segmentation internally in different ways. For models that are trained on sequence level, it may be better if the human reading is also performed the same way. Further research expanding these aspects is desired. Other models beyond GLLMs, \textit{e.g.}~with non-linearity, may be examined, because the probability of regression is within a narrow range in most of the cases. Using models' revisions to predict human behaviour is also a possible research question which was not addressed in this work.

\section*{Acknowledgements}

We thank Nora Hollenstein for some helpful advice on using eye-tracking measures, as well as the authors of the eye-tracking datasets who replied to our clarification requests. Thanks to Patrick Kahardipraja for initial discussions on using surprisal as a signal for revisions policies. We are also thankful for the valuable feedback and suggestions provided by the anonymous reviewers.

\bibliography{anthology,custom}
\bibliographystyle{acl_natbib}

\clearpage
\appendix

\section{Appendix}
\label{sec:appendix}
\section{Pre-processing Human Data}

\begin{table*}[ht!]
\centering
   
    { \setlength{\tabcolsep}{4pt}
    \small
    
    \begin{tabular}{p{2cm}|lp{4cm}|lp{3.5cm}}
        \toprule
        \midrule
         & \textbf{regression} & \textbf{description} & \textbf{skip} & \textbf{description} \\
        \midrule
        \textbf{MECO-L1} and \textbf{MECO-L2} & \texttt{firstrun.reg.out}  & \textit{Variable indicating whether there was a regression from the IA during first-pass reading} &  \texttt{firstrun.skip} & \textit{Variable indicating whether the IA was skipped during first-pass reading}  \\ \midrule
        \textbf{Nicenboim} & \texttt{fp\_reg} & no description & \texttt{FPRT}  & no description \\ \midrule
        \textbf{PoTeC} & \texttt{FPReg} &  \textit{1 if a regression was initiated in the first-pass reading of the word, otherwise 0 (sign(RPD exc))} & negation of \texttt{FPF}  & \textit{1 if the word was fixated in the first-pass, otherwise 0} \\ \midrule
        \textbf{Provo}  & \texttt{IA\_REGRESSION\_OUT} & \textit{Whether the current interest area received at least one regression from later interest areas (e.g., later parts of the sentence). 1 if interest area was entered from a higher IA\_ID (from the right in English); 0 if not. (...) Note that IA\_REGRESSION\_OUT only considers first-pass regressions.} & \texttt{IA\_SKIP}  & \textit{An interest area is considered skipped (i.e., IA\_SKIP = 1) if no fixation occurred in first-pass reading.} \\ \midrule
        \textbf{RastrOS} & \texttt{IA\_REGRESSION\_OUT} & \textit{Whether regression(s) was made from the current interest area to earlier interest areas (e.g., previous parts of the sentence) prior to leaving that interest area in a forward direction. 1 if a saccade exits the current interest area to a lower IA\_ID (to the left in English) before a later interest area was fixated; 0 if not. (...) Note that IA\_REGRESSION\_OUT only considers first-pass regressions.} & \texttt{IA\_SKIP} & \textit{An interest area is considered skipped (i.e.,IA\_SKIP = 1) if no fixation occurred in first-pass reading.} \\ \midrule
        \bottomrule
    \end{tabular}
}
\caption{Measures used for each eye-tracking corpus and their definition according to the available documentation.}
\label{tab:rt-measures}
\end{table*}

We pre-process all datasets to combine the measures into a common format, with one token per row and one column for each subject. If no data was available for a subject, the cell is filled with a NaN value, so that it is later ignored. We partition the measure into three groups: interest areas that were skipped in the first-pass reading (and, consequently, also interest areas that were skipped altogether) are assigned a \texttt{skipped} category (label $-1$). For the remaining interest areas, \textit{i.e.}~those that had a first-pass fixation, we extract either a \texttt{regressed} (label $1$) or \texttt{not regressed} (label $0$) category. Here we document some necessary decisions. The measures we rely on are documented in Table \ref{tab:rt-measures} and the pre-processing scripts are available at \url{https://github.com/briemadu/revreg}. For further details about the data collections, please refer to the original publications.

\vspace{0.2cm}

\noindent $\triangleright$ \textbf{RastrOS}: Participants read paragraphs, one by one in a random order, from journalistic, literary and popular science sources. There was a yes/no comprehension question after 20 of the paragraphs.
We get the tokens from the columns \texttt{Word} and \texttt{IA\_LABEL}. We solve inconsistencies as follows: if \texttt{Word} contains a comma and \texttt{IA\_LABEL} contains a full stop, we use the former (in accordance to personal communication with the author). If there are mismatches in quotation marks, we also use the former. For other inconsistencies (33 tokens), we use the latter. 

\vspace{0.2cm}

\noindent $\triangleright$ \textbf{PoTeC}: Participants read scientific texts on biology and physics from textbooks. Three multiple-choice comprehension questions were presented after each text in a separate screen. 
We use the negation of \texttt{FPF} as an auxiliary to detect tokens that were skipped in the first pass. The raw text files do not contain punctuation in a straightforward format. We thus only extract commas, and final sentence punctuation is considered to be always a full stop, except for two cases that we noticed were not end of sentences, so a ; was used. We follow the list of 13 subjects ids (in the original script \texttt{mergeFixationsWordFeatures.py}) that were removed due to poor calibration (according to \citet{jager2020deep}) and exclude them from our sample.

\vspace{0.2cm}

\noindent $\triangleright$ \textbf{Provo}: Participants read the texts from various sources in a random order, without any additional task. 
For the tokens, we rely on \texttt{IA\_LABEL}, due to inconsistencies in the \texttt{Word} column. Four tokens do not match the raw texts (apparently due to encoding), so we use the text instead of the \texttt{IA\_LABEL}. 

\vspace{0.2cm}

\noindent $\triangleright$ \textbf{MECO-L1}: Wikipedia style texts, each on a separate screen. After each text, there were four yes/no comprehension questions.  
We could only use the Dutch version, as the other languages had mismatches between the source texts and the interest area column. 

\vspace{0.2cm}

\noindent $\triangleright$ \textbf{MECO-L2}: Texts are from training materials for English tests. Participants answered four yes/no questions after each text. 
5 subjects were excluded due to unexplained repetitions.

\vspace{0.2cm}

\noindent $\triangleright$ \textbf{Nicenboim}: Participants read stimuli (sentences). True/false statements appeared randomly after half of them. We use the filler sentences (as the others had varying conditions across participants). We use \texttt{FPRT}, assuming it is first-pass reading time, to infer first-pass fixations: if it is NaN, we consider it to be a skip (because otherwise it is always a number higher than 0).

\section{Pre-processing Models' Data}

We use off-the-shelf implementations of sequence labelling models. To extract the outputs, we loop over the interest areas for each text in the eye-tracking corpus for the corresponding language. At each time step $t$, a string is created with the interest areas up to position $t$, joined with a blank space. The models output a list of labels, which we take to be the output prefix for that time step. Due to the internal tokenization, it can happen in a few cases that tokenization changes slightly or that more than one new label is added. We use the number of labels in the previous time step as a reference, all new labels beyond that length are considered an addition and do not affect revisions. A revision happens if the output prefix at time $t$ differs from the output at time $t-1$; and it is effective if the number of labels that match the final output labels up to that time step increased. For Stanza BiLSTM, we extract the labels from the attributes \texttt{upos\_}, \texttt{deprel}, \texttt{head}. For Explosion's Transformers, we extract the labels from the attributes \texttt{pos\_}, \texttt{dep\_}, \texttt{head\_i}.

\section{Modelling Details}
We fit generalized linear mixed models using the lme4~\cite{lme4} package in the R statistical computing environment~\cite{r}. All baseline and full models were initially fit with the same structure described in the Methods section. We made changes to the model structure in 6 cases to tackle with convergence issues: Model fits to the Nicenboim-TRF-Pos and Nicenboim-BiLSTM-Pos datasets revealed low text-level variance and random effects were excluded in these datasets in further analyses. Token position was not a significant predictor of model revision in MECOL1-BiLSTM-Head, MECOL2-TRF-Head, MECOL2-BiLSTM-Head, and Provo-BiLSTM-Pos models, thus, we refitted these models without the token positions variable. 

\section{Detailed Results}

Tables \ref{tab:results-overview-a} and \ref{tab:results-overview-b} show all the estimated coefficients, standard errors, $z$ and $p$-values for all models. Table \ref{tab:lrt} presents the results of the likelihood ratio tests for the full models in relation to their corresponding null model. All results in the paper have been rounded to to decimal places programatically.

\begin{table*}[ht!]
\centering
   
    { \setlength{\tabcolsep}{5pt}
    \scriptsize
    
\begin{tabular}{llllrrrrrrrr}
\toprule
\midrule
 &  &  &  & \multicolumn{2}{c}{\textbf{estimate}} & \multicolumn{2}{c}{\textbf{SE}} & \multicolumn{2}{c}{\textbf{z}} & \multicolumn{2}{c}{\textbf{p}} \\
 \cmidrule(lr){5-6}\cmidrule(lr){7-8}\cmidrule(lr){9-10}\cmidrule(lr){11-12}
 &  &  &  & \multicolumn{1}{c}{all-r} & \multicolumn{1}{c}{eff-r} & \multicolumn{1}{c}{all-r} & \multicolumn{1}{c}{eff-r} & \multicolumn{1}{c}{all-r} & \multicolumn{1}{c}{eff-r} & \multicolumn{1}{c}{all-r} & \multicolumn{1}{c}{eff-r} \\
\cmidrule(r){1-4}\cmidrule(lr){5-6}\cmidrule(lr){7-8}\cmidrule(lr){9-10}\cmidrule(lr){11-12}
\multirow[t]{24}{*}{\textbf{MECO-L1 (du)}} & \multirow[t]{12}{*}{BiLSTM} & \multirow[t]{4}{*}{deprel} & intercept & 1.47*** & 1.52*** & 0.08 & 0.07 & 17.33 & 23.32 & <0.001 & <0.001 \\
 &  &  & p(reg) & 2.13*** & 1.60*** & 0.12 & 0.11 & 17.04 & 13.97 & <0.001 & <0.001 \\
 &  &  & p(skip) & -2.71*** & -3.48*** & 0.05 & 0.05 & -52.0 & -67.81 & <0.001 & <0.001 \\
 &  &  & position & 0.03** & 0.0 & 0.01 & 0.01 & 3.10 & 0.49 & 0.002 & 0.622 \\
\cmidrule(r){3-4}\cmidrule(lr){5-6}\cmidrule(lr){7-8}\cmidrule(lr){9-10}\cmidrule(lr){11-12}
 &  & \multirow[t]{4}{*}{head} & intercept & 3.34*** & 1.98*** & 0.08 & 0.08 & 40.29 & 25.40 & <0.001 & <0.001 \\
 &  &  & p(reg) & 1.34*** & 1.31*** & 0.16 & 0.11 & 8.51 & 11.47 & <0.001 & <0.001 \\
 &  &  & p(skip) & -4.93*** & -5.14*** & 0.07 & 0.06 & -75.22 & -91.52 & <0.001 & <0.001 \\
 &  &  & position & - & - & - & - & - & - & - & - \\
\cmidrule(r){3-4}\cmidrule(lr){5-6}\cmidrule(lr){7-8}\cmidrule(lr){9-10}\cmidrule(lr){11-12}
 &  & \multirow[t]{4}{*}{pos} & intercept & 0.07 & 0.02 & 0.08 & 0.08 & 0.86 & 0.25 & 0.388 & 0.805 \\
 &  &  & p(reg) & 0.59*** & 0.63*** & 0.12 & 0.13 & 4.77 & 4.89 & <0.001 & <0.001 \\
 &  &  & p(skip) & -2.72*** & -2.96*** & 0.07 & 0.07 & -41.33 & -42.59 & <0.001 & <0.001 \\
 &  &  & position & -0.18*** & -0.18*** & 0.01 & 0.01 & -15.50 & -14.60 & <0.001 & <0.001 \\
\cmidrule(r){2-4}\cmidrule(lr){5-6}\cmidrule(lr){7-8}\cmidrule(lr){9-10}\cmidrule(lr){11-12}
 & \multirow[t]{12}{*}{Transformer} & \multirow[t]{4}{*}{deprel} & intercept & 1.70*** & 1.13*** & 0.09 & 0.07 & 19.25 & 16.35 & <0.001 & <0.001 \\
 &  &  & p(reg) & 2.97*** & 2.38*** & 0.15 & 0.12 & 20.42 & 19.60 & <0.001 & <0.001 \\
 &  &  & p(skip) & -3.08*** & -2.87*** & 0.06 & 0.05 & -54.30 & -56.21 & <0.001 & <0.001 \\
 &  &  & position & 0.07*** & 0.05*** & 0.01 & 0.01 & 7.31 & 6.11 & <0.001 & <0.001 \\
\cmidrule(r){3-4}\cmidrule(lr){5-6}\cmidrule(lr){7-8}\cmidrule(lr){9-10}\cmidrule(lr){11-12}
 &  & \multirow[t]{4}{*}{head} & intercept & 2.17*** & 1.67*** & 0.09 & 0.08 & 23.66 & 21.33 & <0.001 & <0.001 \\
 &  &  & p(reg) & 2.18*** & 1.16*** & 0.16 & 0.11 & 13.87 & 10.48 & <0.001 & <0.001 \\
 &  &  & p(skip) & -3.99*** & -4.49*** & 0.06 & 0.05 & -63.76 & -83.52 & <0.001 & <0.001 \\
 &  &  & position & 0.15*** & -0.0 & 0.01 & 0.01 & 15.93 & -0.27 & <0.001 & 0.789 \\
\cmidrule(r){3-4}\cmidrule(lr){5-6}\cmidrule(lr){7-8}\cmidrule(lr){9-10}\cmidrule(lr){11-12}
 &  & \multirow[t]{4}{*}{pos} & intercept & -2.11*** & -2.22*** & 0.25 & 0.20 & -8.55 & -11.08 & <0.001 & <0.001 \\
 &  &  & p(reg) & 0.59*** & 0.91*** & 0.16 & 0.19 & 3.65 & 4.86 & <0.001 & <0.001 \\
 &  &  & p(skip) & -0.40*** & -0.66*** & 0.08 & 0.09 & -5.08 & -7.05 & <0.001 & <0.001 \\
 &  &  & position & -0.05*** & -0.10*** & 0.01 & 0.02 & -3.45 & -6.13 & <0.001 & <0.001 \\
\cmidrule(r){1-4}\cmidrule(lr){5-6}\cmidrule(lr){7-8}\cmidrule(lr){9-10}\cmidrule(lr){11-12}
\multirow[t]{24}{*}{\textbf{MECO-L2 (en-l2)}} & \multirow[t]{12}{*}{BiLSTM} & \multirow[t]{4}{*}{deprel} & intercept & 1.29*** & 1.04*** & 0.05 & 0.04 & 24.18 & 26.86 & <0.001 & <0.001 \\
 &  &  & p(reg) & 3.41*** & 3.10*** & 0.05 & 0.04 & 73.39 & 72.32 & <0.001 & <0.001 \\
 &  &  & p(skip) & -2.80*** & -2.80*** & 0.02 & 0.02 & -178.47 & -182.23 & <0.001 & <0.001 \\
 &  &  & position & -0.03*** & -0.03*** & 0.0 & 0.0 & -8.94 & -10.02 & <0.001 & <0.001 \\
\cmidrule(r){3-4}\cmidrule(lr){5-6}\cmidrule(lr){7-8}\cmidrule(lr){9-10}\cmidrule(lr){11-12}
 &  & \multirow[t]{4}{*}{head} & intercept & 1.59*** & 0.72*** & 0.06 & 0.06 & 27.44 & 12.28 & <0.001 & <0.001 \\
 &  &  & p(reg) & 4.32*** & 3.27*** & 0.05 & 0.04 & 81.05 & 85.21 & <0.001 & <0.001 \\
 &  &  & p(skip) & -3.23*** & -4.49*** & 0.02 & 0.02 & -193.35 & -257.08 & <0.001 & <0.001 \\
 &  &  & position & - & - & - & - & - & - & - & - \\
\cmidrule(r){3-4}\cmidrule(lr){5-6}\cmidrule(lr){7-8}\cmidrule(lr){9-10}\cmidrule(lr){11-12}
 &  & \multirow[t]{4}{*}{pos} & intercept & -2.62*** & -2.72*** & 0.07 & 0.06 & -36.21 & -44.03 & <0.001 & <0.001 \\
 &  &  & p(reg) & 1.25*** & 1.18*** & 0.05 & 0.05 & 27.53 & 25.28 & <0.001 & <0.001 \\
 &  &  & p(skip) & -1.16*** & -1.23*** & 0.02 & 0.02 & -52.26 & -53.20 & <0.001 & <0.001 \\
 &  &  & position & 0.20*** & 0.21*** & 0.0 & 0.0 & 42.18 & 42.45 & <0.001 & <0.001 \\
\cmidrule(r){2-4}\cmidrule(lr){5-6}\cmidrule(lr){7-8}\cmidrule(lr){9-10}\cmidrule(lr){11-12}
 & \multirow[t]{12}{*}{Transformer} & \multirow[t]{4}{*}{deprel} & intercept & 1.22*** & 1.17*** & 0.09 & 0.07 & 14.28 & 16.69 & <0.001 & <0.001 \\
 &  &  & p(reg) & 4.39*** & 2.56*** & 0.05 & 0.04 & 82.91 & 60.24 & <0.001 & <0.001 \\
 &  &  & p(skip) & -2.53*** & -2.70*** & 0.02 & 0.02 & -154.71 & -176.92 & <0.001 & <0.001 \\
 &  &  & position & 0.03*** & -0.01*** & 0.0 & 0.0 & 11.37 & -5.27 & <0.001 & <0.001 \\
\cmidrule(r){3-4}\cmidrule(lr){5-6}\cmidrule(lr){7-8}\cmidrule(lr){9-10}\cmidrule(lr){11-12}
 &  & \multirow[t]{4}{*}{head} & intercept & 1.45*** & 0.81*** & 0.08 & 0.05 & 18.13 & 17.05 & <0.001 & <0.001 \\
 &  &  & p(reg) & 4.40*** & 3.11*** & 0.05 & 0.04 & 82.24 & 81.62 & <0.001 & <0.001 \\
 &  &  & p(skip) & -2.64*** & -4.93*** & 0.02 & 0.02 & -160.14 & -270.17 & <0.001 & <0.001 \\
 &  &  & position & - & - & - & - & - & - & - & - \\
\cmidrule(r){3-4}\cmidrule(lr){5-6}\cmidrule(lr){7-8}\cmidrule(lr){9-10}\cmidrule(lr){11-12}
 &  & \multirow[t]{4}{*}{pos} & intercept & -2.64*** & -2.62*** & 0.17 & 0.14 & -15.28 & -18.69 & <0.001 & <0.001 \\
 &  &  & p(reg) & -0.62*** & -1.35*** & 0.06 & 0.08 & -9.49 & -17.23 & <0.001 & <0.001 \\
 &  &  & p(skip) & -0.77*** & -0.40*** & 0.03 & 0.03 & -29.33 & -13.58 & <0.001 & <0.001 \\
 &  &  & position & 0.08*** & 0.01* & 0.01 & 0.01 & 15.56 & 2.10 & <0.001 & 0.035 \\
\cmidrule(r){1-4}\cmidrule(lr){5-6}\cmidrule(lr){7-8}\cmidrule(lr){9-10}\cmidrule(lr){11-12}
\multirow[t]{24}{*}{\textbf{Nicenboim (es)}} & \multirow[t]{12}{*}{BiLSTM} & \multirow[t]{4}{*}{deprel} & intercept & 0.42*** & 0.22** & 0.07 & 0.07 & 5.61 & 3.04 & <0.001 & 0.002 \\
 &  &  & p(reg) & 3.35*** & 4.83*** & 0.18 & 0.18 & 18.17 & 26.59 & <0.001 & <0.001 \\
 &  &  & p(skip) & -3.42*** & -3.32*** & 0.05 & 0.05 & -70.86 & -69.87 & <0.001 & <0.001 \\
 &  &  & position & 0.46*** & 0.32*** & 0.02 & 0.02 & 28.17 & 19.62 & <0.001 & <0.001 \\
\cmidrule(r){3-4}\cmidrule(lr){5-6}\cmidrule(lr){7-8}\cmidrule(lr){9-10}\cmidrule(lr){11-12}
 &  & \multirow[t]{4}{*}{head} & intercept & 0.55*** & 0.90*** & 0.07 & 0.08 & 7.97 & 11.03 & <0.001 & <0.001 \\
 &  &  & p(reg) & 4.37*** & 3.22*** & 0.21 & 0.19 & 20.91 & 16.87 & <0.001 & <0.001 \\
 &  &  & p(skip) & -3.74*** & -6.59*** & 0.05 & 0.06 & -73.13 & -102.23 & <0.001 & <0.001 \\
 &  &  & position & 0.59*** & 0.47*** & 0.02 & 0.02 & 34.30 & 23.23 & <0.001 & <0.001 \\
\cmidrule(r){3-4}\cmidrule(lr){5-6}\cmidrule(lr){7-8}\cmidrule(lr){9-10}\cmidrule(lr){11-12}
 &  & \multirow[t]{4}{*}{pos} & intercept & -4.49*** & -4.82*** & 0.07 & 0.08 & -60.34 & -58.02 & <0.001 & <0.001 \\
 &  &  & p(reg) & 6.40*** & 6.58*** & 0.22 & 0.23 & 28.83 & 28.66 & <0.001 & <0.001 \\
 &  &  & p(skip) & 0.74*** & 0.39*** & 0.09 & 0.10 & 8.43 & 4.10 & <0.001 & <0.001 \\
  &  &  & position & 0.31*** & 0.43*** & 0.03 & 0.03 & 10.26 & 12.47 & <0.001 & <0.001 \\
\cmidrule(r){2-4}\cmidrule(lr){5-6}\cmidrule(lr){7-8}\cmidrule(lr){9-10}\cmidrule(lr){11-12}
 & \multirow[t]{12}{*}{Transformer} & \multirow[t]{4}{*}{deprel} & intercept & 0.18* & 0.07 & 0.08 & 0.07 & 2.34 & 0.99 & 0.019 & 0.32 \\
 &  &  & p(reg) & 1.36*** & 1.89*** & 0.16 & 0.15 & 8.68 & 12.29 & <0.001 & <0.001 \\
 &  &  & p(skip) & -2.77*** & -2.80*** & 0.04 & 0.04 & -62.04 & -62.64 & <0.001 & <0.001 \\
 &  &  & position & 0.37*** & 0.30*** & 0.02 & 0.02 & 24.60 & 19.58 & <0.001 & <0.001 \\
\cmidrule(r){3-4}\cmidrule(lr){5-6}\cmidrule(lr){7-8}\cmidrule(lr){9-10}\cmidrule(lr){11-12}
 &  & \multirow[t]{4}{*}{head} & intercept & 0.55*** & 0.82*** & 0.08 & 0.08 & 6.96 & 10.46 & <0.001 & <0.001 \\
 &  &  & p(reg) & 3.09*** & 3.89*** & 0.19 & 0.18 & 16.32 & 21.24 & <0.001 & <0.001 \\
 &  &  & p(skip) & -3.79*** & -5.90*** & 0.05 & 0.06 & -76.03 & -97.66 & <0.001 & <0.001 \\
 &  &  & position & 0.55*** & 0.30*** & 0.02 & 0.02 & 32.66 & 15.40 & <0.001 & <0.001 \\
\cmidrule(r){3-4}\cmidrule(lr){5-6}\cmidrule(lr){7-8}\cmidrule(lr){9-10}\cmidrule(lr){11-12}
 &  & \multirow[t]{4}{*}{pos} & intercept & -2.87*** & -2.72*** & 0.08 & 0.08 & -34.99 & -32.63 & <0.001 & <0.001 \\
 &  &  & p(reg) & 2.92*** & 2.52*** & 0.41 & 0.45 & 7.07 & 5.54 & <0.001 & <0.001 \\
 &  &  & p(skip) & -0.54*** & -0.76*** & 0.13 & 0.14 & -4.06 & -5.41 & <0.001 & <0.001 \\
  &  &  & position & -0.68*** & -0.79*** & 0.04 & 0.05 & -16.06 & -17.45 & <0.001 & <0.001 \\
\midrule
\bottomrule
\end{tabular}
}
\vspace{-0.2cm}
\caption{Overview of all results (part I).}
\label{tab:results-overview-a}
\end{table*}

\begin{table*}[ht!]
\centering
   
    { \setlength{\tabcolsep}{5pt}
    \scriptsize
    
\begin{tabular}{llllrrrrrrrr}
\toprule
\midrule
 &  &  &  & \multicolumn{2}{c}{\textbf{estimate}} & \multicolumn{2}{c}{\textbf{SE}} & \multicolumn{2}{c}{\textbf{z}} & \multicolumn{2}{c}{\textbf{p}} \\
 \cmidrule(lr){5-6}\cmidrule(lr){7-8}\cmidrule(lr){9-10}\cmidrule(lr){11-12}
 &  &  &  & \multicolumn{1}{c}{all-r} & \multicolumn{1}{c}{eff-r} & \multicolumn{1}{c}{all-r} & \multicolumn{1}{c}{eff-r} & \multicolumn{1}{c}{all-r} & \multicolumn{1}{c}{eff-r} & \multicolumn{1}{c}{all-r} & \multicolumn{1}{c}{eff-r} \\
\cmidrule(r){1-4}\cmidrule(lr){5-6}\cmidrule(lr){7-8}\cmidrule(lr){9-10}\cmidrule(lr){11-12}

\multirow[t]{24}{*}{\textbf{PoTeC (de)}} & \multirow[t]{12}{*}{BiLSTM} & \multirow[t]{4}{*}{deprel} & intercept & 0.45*** & 0.14* & 0.08 & 0.06 & 5.98 & 2.17 & <0.001 & 0.03 \\
 &  &  & p(reg) & 1.64*** & 1.77*** & 0.06 & 0.06 & 25.48 & 29.24 & <0.001 & <0.001 \\
 &  &  & p(skip) & -3.18*** & -2.98*** & 0.04 & 0.04 & -86.61 & -81.10 & <0.001 & <0.001 \\
 &  &  & position & 0.07*** & 0.02*** & 0.01 & 0.01 & 10.20 & 3.51 & <0.001 & <0.001 \\
\cmidrule(r){3-4}\cmidrule(lr){5-6}\cmidrule(lr){7-8}\cmidrule(lr){9-10}\cmidrule(lr){11-12}
 &  & \multirow[t]{4}{*}{head} & intercept & 0.95*** & -0.17*** & 0.07 & 0.05 & 14.36 & -3.59 & <0.001 & <0.001 \\
 &  &  & p(reg) & 3.80*** & 3.99*** & 0.09 & 0.07 & 41.28 & 57.10 & <0.001 & <0.001 \\
 &  &  & p(skip) & -4.16*** & -4.99*** & 0.04 & 0.04 & -100.28 & -113.17 & <0.001 & <0.001 \\
 &  &  & position & 0.12*** & 0.07*** & 0.01 & 0.01 & 15.72 & 9.71 & <0.001 & <0.001 \\
\cmidrule(r){3-4}\cmidrule(lr){5-6}\cmidrule(lr){7-8}\cmidrule(lr){9-10}\cmidrule(lr){11-12}
 &  & \multirow[t]{4}{*}{pos} & intercept & -1.88*** & -1.89*** & 0.07 & 0.07 & -26.25 & -26.58 & <0.001 & <0.001 \\
 &  &  & p(reg) & 2.38*** & 2.66*** & 0.06 & 0.07 & 37.30 & 40.46 & <0.001 & <0.001 \\
 &  &  & p(skip) & -2.63*** & -2.78*** & 0.05 & 0.05 & -52.65 & -51.57 & <0.001 & <0.001 \\
 &  &  & position & 0.12*** & 0.07*** & 0.01 & 0.01 & 13.44 & 7.61 & <0.001 & <0.001 \\
\cmidrule(r){2-4}\cmidrule(lr){5-6}\cmidrule(lr){7-8}\cmidrule(lr){9-10}\cmidrule(lr){11-12}
 & \multirow[t]{12}{*}{Transformer} & \multirow[t]{4}{*}{deprel} & intercept & 0.62*** & 0.28*** & 0.07 & 0.05 & 9.23 & 5.88 & <0.001 & <0.001 \\
 &  &  & p(reg) & 4.53*** & 2.86*** & 0.10 & 0.07 & 45.75 & 41.76 & <0.001 & <0.001 \\
 &  &  & p(skip) & -2.62*** & -2.29*** & 0.04 & 0.04 & -64.82 & -64.33 & <0.001 & <0.001 \\
 &  &  & position & 0.14*** & 0.04*** & 0.01 & 0.01 & 19.63 & 5.69 & <0.001 & <0.001 \\
\cmidrule(r){3-4}\cmidrule(lr){5-6}\cmidrule(lr){7-8}\cmidrule(lr){9-10}\cmidrule(lr){11-12}
 &  & \multirow[t]{4}{*}{head} & intercept & 0.46*** & -0.20*** & 0.06 & 0.05 & 7.90 & -4.53 & <0.001 & <0.001 \\
 &  &  & p(reg) & 5.36*** & 3.31*** & 0.11 & 0.07 & 50.06 & 49.28 & <0.001 & <0.001 \\
 &  &  & p(skip) & -2.63*** & -4.08*** & 0.04 & 0.04 & -63.93 & -101.16 & <0.001 & <0.001 \\
 &  &  & position & 0.17*** & 0.10*** & 0.01 & 0.01 & 23.09 & 13.98 & <0.001 & <0.001 \\
\cmidrule(r){3-4}\cmidrule(lr){5-6}\cmidrule(lr){7-8}\cmidrule(lr){9-10}\cmidrule(lr){11-12}
 &  & \multirow[t]{4}{*}{pos} & intercept & -2.40*** & -2.47*** & 0.13 & 0.11 & -18.18 & -21.86 & <0.001 & <0.001 \\
 &  &  & p(reg) & 1.32*** & 1.12*** & 0.12 & 0.13 & 11.36 & 8.57 & <0.001 & <0.001 \\
 &  &  & p(skip) & 0.32*** & 0.36*** & 0.08 & 0.08 & 4.29 & 4.37 & <0.001 & <0.001 \\
 &  &  & position & -0.23*** & -0.25*** & 0.01 & 0.01 & -18.32 & -18.61 & <0.001 & <0.001 \\
\cmidrule(r){1-4}\cmidrule(lr){5-6}\cmidrule(lr){7-8}\cmidrule(lr){9-10}\cmidrule(lr){11-12}
\multirow[t]{24}{*}{\textbf{Provo (en)}} & \multirow[t]{12}{*}{BiLSTM} & \multirow[t]{4}{*}{deprel} & intercept & 1.22*** & 0.80*** & 0.05 & 0.04 & 24.29 & 18.20 & <0.001 & <0.001 \\
 &  &  & p(reg) & 3.30*** & 2.95*** & 0.09 & 0.08 & 38.56 & 38.54 & <0.001 & <0.001 \\
 &  &  & p(skip) & -3.68*** & -3.66*** & 0.03 & 0.03 & -133.52 & -137.30 & <0.001 & <0.001 \\
 &  &  & position & 0.21*** & 0.24*** & 0.01 & 0.01 & 38.87 & 43.77 & <0.001 & <0.001 \\
\cmidrule(r){3-4}\cmidrule(lr){5-6}\cmidrule(lr){7-8}\cmidrule(lr){9-10}\cmidrule(lr){11-12}
 &  & \multirow[t]{4}{*}{head} & intercept & 1.76*** & 0.41*** & 0.05 & 0.04 & 33.12 & 10.44 & <0.001 & <0.001 \\
 &  &  & p(reg) & 2.18*** & 1.36*** & 0.10 & 0.07 & 21.84 & 20.64 & <0.001 & <0.001 \\
 &  &  & p(skip) & -4.92*** & -4.57*** & 0.03 & 0.03 & -155.18 & -161.06 & <0.001 & <0.001 \\
 &  &  & position & 0.40*** & 0.31*** & 0.01 & 0.01 & 68.85 & 54.05 & <0.001 & <0.001 \\
\cmidrule(r){3-4}\cmidrule(lr){5-6}\cmidrule(lr){7-8}\cmidrule(lr){9-10}\cmidrule(lr){11-12}
 &  & \multirow[t]{4}{*}{pos} & intercept & -1.92*** & -2.02*** & 0.08 & 0.08 & -22.77 & -25.78 & <0.001 & <0.001 \\
 &  &  & p(reg) & 1.42*** & 1.58*** & 0.08 & 0.08 & 18.61 & 20.38 & <0.001 & <0.001 \\
 &  &  & p(skip) & -0.66*** & -0.77*** & 0.04 & 0.04 & -18.63 & -20.72 & <0.001 & <0.001 \\
 &  &  & position & - & - & - & - & - & - & - & - \\
\cmidrule(r){2-4}\cmidrule(lr){5-6}\cmidrule(lr){7-8}\cmidrule(lr){9-10}\cmidrule(lr){11-12}
 & \multirow[t]{12}{*}{Transformer} & \multirow[t]{4}{*}{deprel} & intercept & 1.28*** & 0.93*** & 0.05 & 0.04 & 24.39 & 23.44 & <0.001 & <0.001 \\
 &  &  & p(reg) & 3.26*** & 2.69*** & 0.09 & 0.08 & 34.39 & 33.70 & <0.001 & <0.001 \\
 &  &  & p(skip) & -3.75*** & -3.41*** & 0.03 & 0.03 & -129.34 & -127.32 & <0.001 & <0.001 \\
 &  &  & position & 0.30*** & 0.24*** & 0.01 & 0.01 & 54.95 & 45.93 & <0.001 & <0.001 \\
\cmidrule(r){3-4}\cmidrule(lr){5-6}\cmidrule(lr){7-8}\cmidrule(lr){9-10}\cmidrule(lr){11-12}
 &  & \multirow[t]{4}{*}{head} & intercept & 1.45*** & 0.46*** & 0.05 & 0.04 & 29.17 & 11.59 & <0.001 & <0.001 \\
 &  &  & p(reg) & 2.27*** & 1.69*** & 0.10 & 0.07 & 23.76 & 25.60 & <0.001 & <0.001 \\
 &  &  & p(skip) & -4.01*** & -4.66*** & 0.03 & 0.03 & -133.24 & -163.42 & <0.001 & <0.001 \\
 &  &  & position & 0.37*** & 0.28*** & 0.01 & 0.01 & 64.92 & 48.09 & <0.001 & <0.001 \\
\cmidrule(r){3-4}\cmidrule(lr){5-6}\cmidrule(lr){7-8}\cmidrule(lr){9-10}\cmidrule(lr){11-12}
 &  & \multirow[t]{4}{*}{pos} & intercept & -2.69*** & -2.89*** & 0.14 & 0.13 & -19.71 & -23.06 & <0.001 & <0.001 \\
 &  &  & p(reg) & 3.00*** & 3.15*** & 0.10 & 0.10 & 31.11 & 30.24 & <0.001 & <0.001 \\
 &  &  & p(skip) & 0.80*** & 0.93*** & 0.04 & 0.05 & 18.07 & 19.09 & <0.001 & <0.001 \\
 &  &  & position & -0.25*** & -0.27*** & 0.01 & 0.01 & -30.18 & -29.77 & <0.001 & <0.001 \\
\cmidrule(r){1-4}\cmidrule(lr){5-6}\cmidrule(lr){7-8}\cmidrule(lr){9-10}\cmidrule(lr){11-12}
\multirow[t]{24}{*}{\textbf{RastrOS (pt-br)}} & \multirow[t]{12}{*}{BiLSTM} & \multirow[t]{4}{*}{deprel} & intercept & -0.22*** & -0.32*** & 0.05 & 0.05 & -4.68 & -7.01 & <0.001 & <0.001 \\
 &  &  & p(reg) & 4.16*** & 3.62*** & 0.08 & 0.07 & 51.32 & 49.69 & <0.001 & <0.001 \\
 &  &  & p(skip) & -1.70*** & -2.05*** & 0.03 & 0.03 & -56.48 & -65.25 & <0.001 & <0.001 \\
 &  &  & position & 0.14*** & 0.12*** & 0.01 & 0.01 & 17.09 & 14.20 & <0.001 & <0.001 \\
\cmidrule(r){3-4}\cmidrule(lr){5-6}\cmidrule(lr){7-8}\cmidrule(lr){9-10}\cmidrule(lr){11-12}
 &  & \multirow[t]{4}{*}{head} & intercept & -0.15** & -0.19*** & 0.05 & 0.05 & -2.90 & -3.69 & 0.004 & <0.001 \\
 &  &  & p(reg) & 4.66*** & 4.03*** & 0.10 & 0.08 & 48.70 & 50.46 & <0.001 & <0.001 \\
 &  &  & p(skip) & -2.17*** & -4.13*** & 0.03 & 0.04 & -69.37 & -102.76 & <0.001 & <0.001 \\
 &  &  & position & 0.26*** & 0.19*** & 0.01 & 0.01 & 30.43 & 20.73 & <0.001 & <0.001 \\
\cmidrule(r){3-4}\cmidrule(lr){5-6}\cmidrule(lr){7-8}\cmidrule(lr){9-10}\cmidrule(lr){11-12}
 &  & \multirow[t]{4}{*}{pos} & intercept & -0.99*** & -0.98*** & 0.06 & 0.06 & -15.45 & -16.45 & <0.001 & <0.001 \\
 &  &  & p(reg) & 2.63*** & 2.61*** & 0.06 & 0.06 & 43.02 & 42.69 & <0.001 & <0.001 \\
 &  &  & p(skip) & -2.52*** & -2.91*** & 0.04 & 0.04 & -65.44 & -69.72 & <0.001 & <0.001 \\
 &  &  & position & 0.12*** & 0.11*** & 0.01 & 0.01 & 13.14 & 11.68 & <0.001 & <0.001 \\
\cmidrule(r){2-4}\cmidrule(lr){5-6}\cmidrule(lr){7-8}\cmidrule(lr){9-10}\cmidrule(lr){11-12}
 & \multirow[t]{12}{*}{Transformer} & \multirow[t]{4}{*}{deprel} & intercept & -0.10 & -0.26*** & 0.06 & 0.05 & -1.79 & -5.77 & 0.073 & <0.001 \\
 &  &  & p(reg) & 3.12*** & 3.07*** & 0.07 & 0.07 & 42.88 & 45.31 & <0.001 & <0.001 \\
 &  &  & p(skip) & -1.78*** & -2.06*** & 0.03 & 0.03 & -59.32 & -65.27 & <0.001 & <0.001 \\
 &  &  & position & 0.13*** & 0.08*** & 0.01 & 0.01 & 16.10 & 10.07 & <0.001 & <0.001 \\
\cmidrule(r){3-4}\cmidrule(lr){5-6}\cmidrule(lr){7-8}\cmidrule(lr){9-10}\cmidrule(lr){11-12}
 &  & \multirow[t]{4}{*}{head} & intercept & -0.07 & -0.26*** & 0.06 & 0.05 & -1.14 & -4.81 & 0.255 & <0.001 \\
 &  &  & p(reg) & 3.95*** & 3.74*** & 0.09 & 0.08 & 43.65 & 48.84 & <0.001 & <0.001 \\
 &  &  & p(skip) & -2.17*** & -3.93*** & 0.03 & 0.04 & -69.58 & -100.10 & <0.001 & <0.001 \\
 &  &  & position & 0.28*** & 0.20*** & 0.01 & 0.01 & 31.87 & 21.72 & <0.001 & <0.001 \\
\cmidrule(r){3-4}\cmidrule(lr){5-6}\cmidrule(lr){7-8}\cmidrule(lr){9-10}\cmidrule(lr){11-12}
 &  & \multirow[t]{4}{*}{pos} & intercept & -1.97*** & -2.12*** & 0.14 & 0.12 & -14.37 & -17.71 & <0.001 & <0.001 \\
 &  &  & p(reg) & 0.57*** & 0.78*** & 0.09 & 0.09 & 6.32 & 8.43 & <0.001 & <0.001 \\
 &  &  & p(skip) & -0.36*** & -0.65*** & 0.05 & 0.06 & -7.14 & -11.52 & <0.001 & <0.001 \\
 &  &  & position & -0.21*** & -0.18*** & 0.01 & 0.01 & -16.20 & -13.25 & <0.001 & <0.001 \\
\midrule
\bottomrule
\end{tabular}
}
\vspace{-0.2cm}
\caption{Overview of all results (part 2).}
\label{tab:results-overview-b}
\end{table*}

\begin{table*}[ht!]
\centering
   
    { \setlength{\tabcolsep}{5pt}
    \footnotesize

\begin{tabular}{lllrrrrcccc}
\toprule
\midrule
 &  &  & \multicolumn{2}{c}{\textbf{BIC}} & \multicolumn{2}{c}{\textbf{$\chi^2$}} & \multicolumn{2}{c}{\textbf{Df}} & \multicolumn{2}{c}{\textbf{p}} \\
 \cmidrule(lr){4-5}\cmidrule(lr){6-7}\cmidrule(lr){8-9}\cmidrule(lr){10-11}
 &  & & \multicolumn{1}{c}{all-r} & \multicolumn{1}{c}{eff-r} & \multicolumn{1}{c}{all-r} & \multicolumn{1}{c}{eff-r} & \multicolumn{1}{c}{all-r} & \multicolumn{1}{c}{eff-r} & \multicolumn{1}{c}{all-r} & \multicolumn{1}{c}{eff-r} \\
 \cmidrule(r){1-3}\cmidrule(lr){4-5}\cmidrule(lr){6-7}\cmidrule(lr){8-9}\cmidrule(lr){10-11}
\multirow[t]{6}{*}{\textbf{MECO-L1 (du)}} & \multirow[t]{3}{*}{BiLSTM} & deprel & 83546.50 & 82400.75 & 7670.49 & 11010.84 & 2 & 2 & <0.001 & <0.001 \\
 &  & head & 72210.57 & 71300.38 & 14407.59 & 18647.46 & 2 & 2 & <0.001 & <0.001 \\
 &  & pos & 48702.71 & 44738.04 & 3086.69 & 3257.63 & 2 & 2 & <0.001 & <0.001 \\
 \cmidrule(r){2-3}\cmidrule(lr){4-5}\cmidrule(lr){6-7}\cmidrule(lr){8-9}\cmidrule(lr){10-11}
 & \multirow[t]{3}{*}{Transformer} & deprel & 78422.59 & 84143.92 & 9326.01 & 9114.93 & 2 & 2 & <0.001 & <0.001 \\
 &  & head & 73396.78 & 74729.62 & 11051.14 & 15030.07 & 2 & 2 & <0.001 & <0.001 \\
 &  & pos & 40292.37 & 30157.89 & 104.20 & 188.25 & 2 & 2 & <0.001 & <0.001 \\
 \cmidrule(r){1-3}\cmidrule(lr){4-5}\cmidrule(lr){6-7}\cmidrule(lr){8-9}\cmidrule(lr){10-11}
\multirow[t]{6}{*}{\textbf{MECO-L2 (en-l2)}} & \multirow[t]{3}{*}{BiLSTM} & deprel & 786910.36 & 813266.48 & 80770.80 & 80710.35 & 2 & 2 & <0.001 & <0.001 \\
 &  & head & 721580.34 & 719579.16 & 99023.63 & 143061.55 & 2 & 2 & <0.001 & <0.001 \\
 &  & pos & 453665.09 & 428746.07 & 6319.52 & 6111.91 & 2 & 2 & <0.001 & <0.001 \\
 \cmidrule(r){2-3}\cmidrule(lr){4-5}\cmidrule(lr){6-7}\cmidrule(lr){8-9}\cmidrule(lr){10-11}
 & \multirow[t]{3}{*}{Transformer} & deprel & 733070.61 & 810697.01 & 71546.51 & 69594.78 & 2 & 2 & <0.001 & <0.001 \\
 &  & head & 721451.14 & 700433.46 & 74786.38 & 156206.75 & 2 & 2 & <0.001 & <0.001 \\
 &  & pos & 339878.15 & 289129.13 & 891.56 & 320.93 & 2 & 2 & <0.001 & <0.001 \\
 \cmidrule(r){1-3}\cmidrule(lr){4-5}\cmidrule(lr){6-7}\cmidrule(lr){8-9}\cmidrule(lr){10-11}
\multirow[t]{6}{*}{\textbf{Nicenboim (es)}} & \multirow[t]{3}{*}{BiLSTM} & deprel & 62038.56 & 61705.57 & 12760.80 & 14065.35 & 2 & 2 & <0.001 & <0.001 \\
 &  & head & 57808.46 & 47871.65 & 14703.78 & 28111.09 & 2 & 2 & <0.001 & <0.001 \\
 &  & pos & - & - & - & - & 2 & 2 & <0.001 & <0.001 \\
 \cmidrule(r){2-3}\cmidrule(lr){4-5}\cmidrule(lr){6-7}\cmidrule(lr){8-9}\cmidrule(lr){10-11}
 & \multirow[t]{3}{*}{Transformer} & deprel & 67874.18 & 67156.03 & 7930.09 & 8486.24 & 2 & 2 & <0.001 & <0.001 \\
 &  & head & 59810.38 & 50487.63 & 14257.23 & 25095.62 & 2 & 2 & <0.001 & <0.001 \\
 &  & pos & - & - & - & - & 2 & 2 & <0.001 & <0.001 \\
 \cmidrule(r){1-3}\cmidrule(lr){4-5}\cmidrule(lr){6-7}\cmidrule(lr){8-9}\cmidrule(lr){10-11}
\multirow[t]{6}{*}{\textbf{Potec (de)}} & \multirow[t]{3}{*}{BiLSTM} & deprel & 145892.59 & 146622.24 & 15375.96 & 14146.19 & 2 & 2 & <0.001 & <0.001 \\
 &  & head & 121763.91 & 123406.25 & 26247.07 & 35086.70 & 2 & 2 & <0.001 & <0.001 \\
 &  & pos & 101606.89 & 92595.02 & 8237.35 & 8481.05 & 2 & 2 & <0.001 & <0.001 \\
 \cmidrule(r){2-3}\cmidrule(lr){4-5}\cmidrule(lr){6-7}\cmidrule(lr){8-9}\cmidrule(lr){10-11}
 & \multirow[t]{3}{*}{Transformer} & deprel & 119666.13 & 148027.04 & 15136.50 & 12789.51 & 2 & 2 & <0.001 & <0.001 \\
 &  & head & 116103.04 & 133947.43 & 16464.67 & 26794.85 & 2 & 2 & <0.001 & <0.001 \\
 &  & pos & 45668.14 & 39611.90 & 122.34 & 69.33 & 2 & 2 & <0.001 & <0.001 \\
 \cmidrule(r){1-3}\cmidrule(lr){4-5}\cmidrule(lr){6-7}\cmidrule(lr){8-9}\cmidrule(lr){10-11}
\multirow[t]{6}{*}{\textbf{Provo (en)}} & \multirow[t]{3}{*}{BiLSTM} & deprel & 265555.00 & 274893.44 & 38215.03 & 39111.65 & 2 & 2 & <0.001 & <0.001 \\
 &  & head & 235978.14 & 258861.40 & 47478.23 & 46669.46 & 2 & 2 & <0.001 & <0.001 \\
 &  & pos & 166971.65 & 155560.74 & 1373.30 & 1651.36 & 2 & 2 & <0.001 & <0.001 \\
 \cmidrule(r){2-3}\cmidrule(lr){4-5}\cmidrule(lr){6-7}\cmidrule(lr){8-9}\cmidrule(lr){10-11}
 & \multirow[t]{3}{*}{Transformer} & deprel & 252130.08 & 275598.83 & 35534.99 & 33376.38 & 2 & 2 & <0.001 & <0.001 \\
 &  & head & 243848.75 & 255250.75 & 34344.11 & 49079.47 & 2 & 2 & <0.001 & <0.001 \\
 &  & pos & 118204.33 & 103652.19 & 886.80 & 843.64 & 2 & 2 & <0.001 & <0.001 \\
 \cmidrule(r){1-3}\cmidrule(lr){4-5}\cmidrule(lr){6-7}\cmidrule(lr){8-9}\cmidrule(lr){10-11}
\multirow[t]{6}{*}{\textbf{RastrOS (pt-br)}} & \multirow[t]{3}{*}{BiLSTM} & deprel & 106976.51 & 106122.58 & 13333.03 & 14659.62 & 2 & 2 & <0.001 & <0.001 \\
 &  & head & 99489.29 & 89499.43 & 16638.70 & 30213.54 & 2 & 2 & <0.001 & <0.001 \\
 &  & pos & 92050.16 & 87854.00 & 12436.14 & 13744.63 & 2 & 2 & <0.001 & <0.001 \\
 \cmidrule(r){2-3}\cmidrule(lr){4-5}\cmidrule(lr){6-7}\cmidrule(lr){8-9}\cmidrule(lr){10-11}
 & \multirow[t]{3}{*}{Transformer} & deprel & 108432.79 & 106773.67 & 11382.04 & 13219.43 & 2 & 2 & <0.001 & <0.001 \\
 &  & head & 99987.62 & 91261.27 & 15027.19 & 27893.16 & 2 & 2 & <0.001 & <0.001 \\
 &  & pos & 49397.83 & 44548.61 & 169.15 & 371.72 & 2 & 2 & <0.001 & <0.001 \\
\midrule
\bottomrule
\end{tabular}

}
\caption{Overview of likelihood ratio tests, showing how each full model compares to the null model.}
\label{tab:lrt}
\end{table*}

\end{document}